\documentclass[sigconf]{acmart}
\copyrightyear{2020} 
\acmYear{2020} 
\setcopyright{none}
\acmConference[GECCO '20]{Genetic and Evolutionary Computation Conference}{July 8--12, 2020}{Cancún, Mexico}
\acmBooktitle{Genetic and Evolutionary Computation Conference (GECCO '20), July 8--12, 2020, Cancún, Mexico}
\acmPrice{15.00}
\acmDOI{10.1145/3377930.3390155}
\acmISBN{978-1-4503-7128-5/20/07}

\usepackage{amsmath,amssymb,amsthm}
\usepackage{xspace}
\usepackage{epsfig}
\usepackage{enumitem}
\usepackage{algorithm}
\usepackage[noend]{algorithmic}
\usepackage{color}
\usepackage{url}
\usepackage{booktabs,multirow}
\usepackage{tikz}
\usepackage{verbatim}
\usetikzlibrary{arrows,shapes}

\renewcommand{\epsilon}{\varepsilon}
\newcommand{\R}{\mathbb{R}}

\newcommand{\M}{\mathbb{M}}
\newcommand{\Er}{\mathcal{R}}

\newcommand{\RmlrMBO}{\texttt{mlrMBO}\xspace}
\newcommand{\Rsmoof}{\texttt{smoof}\xspace}
\newcommand{\Rqrng}{\texttt{qrng}\xspace}
\newcommand{\Rrandtoolbox}{\texttt{randtoolbox}\xspace}
\newcommand{\Rlhs}{\texttt{lhs}\xspace}
\newcommand{\Rdandy}{\texttt{dandy}\xspace}

\DeclareMathOperator{\VBS}{VBS}

\begin{document}


% \ccsdesc[100]{Theory of computation~Online learning algorithms}

% \begin{CCSXML}
% <ccs2012>
% <concept>
% <concept_id>10003752.10010070.10011796</concept_id>
% <concept_desc>Theory of computation~Theory of randomized search~heuristics</concept_desc>
% <concept_significance>500</concept_significance>
% </concept>
% </ccs2012>
% \end{CCSXML}

% \ccsdesc[500]{Theory of computation ~ Theory of randomized search heuristics}


\author{Jakob Bossek}
\affiliation{%
  \institution{School of Computer Science}
  \city{The University of Adelaide, Australia}
}

\author{Carola Doerr}
\affiliation{%
  \institution{Sorbonne University, CNRS, LIP6}
  \city{Paris, France}
}

\author{Pascal Kerschke}
\affiliation{%
  \institution{Information Systems and Statistics}
  \city{University of M{\"u}nster, Germany}
}

\title[Effects of Initial Design Strategies on SMBO Performance]{Initial Design Strategies and their Effects on\\ Sequential Model-Based Optimization}
\subtitle{An Exploratory Case Study Based on BBOB}

\begin{abstract}
Sequential model-based optimization (SMBO) approaches are algorithms for solving problems that require computationally or otherwise expensive function evaluations. The key design principle of SMBO is a substitution of the true objective function by a surrogate, which is used to propose the point(s) to be evaluated next.  

SMBO algorithms are intrinsically modular, leaving the user with many important design choices. Significant research efforts go into understanding which settings perform best for which type of problems. Most works, however, focus on the choice of the model, the acquisition function, and the strategy used to optimize the latter. The choice of the initial sampling strategy, however, receives much less attention. Not surprisingly, quite diverging recommendations can be found in the literature. 

We analyze in this work how the size and the distribution of the initial sample influences the overall quality of the efficient global optimization~(EGO) algorithm, a well-known SMBO approach. While, overall, small initial budgets using Halton sampling seem preferable, we also observe that the performance landscape is rather unstructured. We furthermore identify several situations in which EGO performs unfavorably against random sampling. Both observations indicate that an adaptive SMBO design could be beneficial, making SMBO an interesting test-bed for automated algorithm design.
\end{abstract}

\maketitle

\sloppy{
\section{Introduction}
\label{sec:intro}

\emph{Sequential Model-Based Optimization} (SMBO) algorithms are techniques for the optimization of problems for which the evaluation of solution candidates is resource-intensive, such as problems requiring real physical experiments or problems that require computationally-expensive simulations. The latter are particularly present in almost any application of Artificial Intelligence, most notably in terms of parameter tuning problems -- a problem that is also omnipresent in Evolutionary Computation~\cite{LoboLM07}. SMBO-based techniques are among the most successfully applied hyper-parameter tuning methods~\cite{SMAC,SPOT,BOHB,KotthoffTHHL19}, so that research on this family of iterative optimization heuristics has gained significant traction in the last decade. SMBO forms today an integral part of the state-of-the-art heuristic solvers. Its probably best-known representatives are \emph{Bayesian Optimization}~(see surveys by \cite{ShahriariSWAF16,MockusBO,GaussianPML}) and, in particular, \emph{Efficient Global Optimization} (EGO,~\cite{JonesSW98}). 

The generic SMBO method works as follows. An initial design of points is sampled and evaluated with the true objective function. The eponymous \emph{sequential} part iteratively (1) builds a surrogate of the true objective function (on basis of the already evaluated samples), (2) proposes new samples by optimizing a so-called \emph{infill-criterion} (which is sometimes referred to as \emph{acquisition function}), (3)~evaluates these additional samples, and (4) integrates these samples, together with their quality indicators (``function values'', ``fitness'') into the memory. 
Each of these steps offers a great variety of design choices, which all may affect the performance of the SMBO procedure. Which surrogate model should be used? Which of the countless infill criteria to use? What method should be used to create the initial sample and what proportion of the overall budget should be spent on the initial design? 
While a large body of works addresses the first two questions (see surveys mentioned above), the latter two questions are treated rather poorly. 
In this work we aim to shed light on the relevance of a suitably chosen initial sampling strategy. More precisely, we study how the size of the initial design and the strategy used to generate it affects the performance of SMBO. As a well-established benchmark environment offering a great variety of different numerical optimization problems, we chose the 24 noiseless BBOB functions (in different dimensions) as test-bed for our investigation.

Our setup comprises of varying the initial design strategy (classical uniform and Latin-Hypercube-Sampling (LHS) as the most frequently used methods and quasi-random Halton and Sobol' sequences), the total budget, and the fraction of this total budget that is used to build the initial sample. We study a total of 720 problems, which are evaluated against 40 different initial design strategies. 

Our general observation is that SMBO performance tends to decrease with increasing initial design ratio, which is in line with the general expectation that adaptive search should outperform non-adaptive sampling. This may justify extreme settings such as the singleton initial design used in the SMAC parameter tuning framework~\cite{SMAC}. 
As always in simulation-based optimization, we are confronted with the important trade-off between the exploitation of already acquired knowledge (through adaptive sampling) and the reduction of uncertainty in regions of the search space that are currently not well covered with already evaluated samples. Sampling in the latter regions of high uncertainty -- commonly referred to as exploration -- can help to identify other promising regions of the search space. In our experiments, we observe indeed that small initial designs are not always preferable. In fact, we even identify cases in which pure (quasi-)random sampling outperforms any of the tested SMBO-based techniques. 

We use our huge database also to investigate advantages of long runs vs. restarted ones. That is, we address the question whether one should use the full budget for one long run, or whether two shorter runs of smaller budget are preferable. We identify several cases in which restarts seem preferable, giving another indication that an adaptive design of SMBO techniques could be preferable. 

The evaluation and analysis of the dataset (which comprises more than $500\,000$ experiments) has been particularly challenging, as no clear pattern between the performance of the different designs and the parameters of the problem (such as its dimension, its high-level features, or even its function ID) were observable. Our data suggests that machine-trained algorithm configuration techniques should be able to outperform state-of-the-art SMBO designs by large margins. The appropriateness of the BBOB dataset for finding generalizable patterns has been shown in~\cite{BelkhirDSS17,KerschkeT2019AutomatedAlgorithm}.

\vspace*{-0.025cm}
\paragraph{Paper Organization.}
This work is structured as follows. Below, we continue with an overview of related work and give information about the availability of our data. Section~\ref{sec:smbo} details the SMBO approach. In Section~\ref{sec:expsetup} we describe our experimental setup including considered benchmark problems, parameter choices and performance measures. Results are presented in Sections~\ref{sec:results-overall} to \ref{sec:restart}. We conclude with final remarks and visions for incorporating the acquired knowledge into improved SMBO approaches.

\vspace*{-0.025cm}
\paragraph{Related Work}
For surveys on Bayesian optimization and, more generally, SMBO approaches we refer the interested reader to the already mentioned surveys~\cite{ShahriariSWAF16,MockusBO,GaussianPML}. Our work builds on EGO, originally suggested by Jones, Schonlau, and Welch~\cite{JonesSW98}. EGO is characterized by using a flexible Kriging, i.e., a Gaussian process surrogate model which offers a natural uncertainty estimate and the widely used quasi-standard expected improvement (EI) infill criterion which balances exploitation of the model and exploration of uncertain regions of the model~\cite{Jones2001}.

Our key interest is an analysis of the influence of the initial design's size and distribution. We assess four different distributions: uniform sampling, LHS, Halton points, and Sobol' sequences. For each of these designs we test ten different initial sample sizes. 
Recommendations on which initial design should be favored vary quite significantly within the community, see~\cite{MorarKS17,BartzP06} for a discussion. In terms of design \textit{size}, 
SMAC~\cite{SMAC} makes an extreme choice in that it uses only one randomly sampled initial design point, whereas other commonly found SMBO implementations typically operate with an initial design of size $10 \cdot d$~\cite{JonesSW98,MorarKS17}, where $d$ denotes the search space dimension (i.e., the optimization problem can be modeled as a function $f:S \subseteq \R^d \to \R$). 
In terms of design \emph{distribution}, LHS and uniform sampling are routinely used in SMBO applications, while quasi-random designs, like Halton and Sobol' designs, are less commonly found -- despite several indications that their even distribution may be beneficial for maximizing the initial exploration~\cite{santner_design_2003}.

We next summarize the main works which explicitly address the question how to chose the initial design. 

Bartz-Beielstein and Preuss study in~\cite{BartzP06} suitable initial designs for SPOT~\cite{SPOT}, an SMBO algorithm specifically designed to perform well on parameter tuning challenges. From experiments on hyperparameter tuning of evolutionary computation techniques, they conclude that LHS sampling is, in general, to be preferred over uniform sampling. They thereby disagree with statements previously made in~\cite{santner_design_2003}, which argues that LHS designs do not gain much over uniform sampling, and that quasi-random sampling strategies should be used instead. The recommendation in~\cite{santner_design_2003} is, however, to be understood in terms of general design of experiments setting, and not specifically addressing SMBO initialization.

Brockhoff et al.~\cite{BrockhoffBW15} studied the difference between random sampling and LHS designs for Matlab's MATSuMoTo model-based optimizer~\cite{mueller2014matsumoto}. In contrast to our work, they fix the total budget of function evaluations to $n=50 \cdot d$ (whereas we use $n = 2^4, \ldots, 2^9$) and compared only four initial designs: LHS with $2\cdot(d+1)\cdot k$ for $k=1,2,10$ and random sampling with $4 \cdot (d+1)$ points. Results are compared against SMAC~\cite{SMAC} and pure random sampling. Their experiments are also across all 24 BBOB functions in $d=2,3,5,10,20$ dimensions (we study $d=2,3,4,5,10$). Their performance measure is a fixed-target measure, more precisely they study the expected running time (ERT) for target values that are chosen individually for each function and they also compare the anytime performance in terms of ECDF curves. Based on their experiments, Brockhoff et al. conclude that for this setting, no clear advantage of LHS designs can be observed and that large initial samples seem detrimental.

Morar et al.~\cite{MorarKS17} also compare LHS and uniform sampling, but fix the size of the initial design to $2 \cdot d$ and rather focus on the interplay between initial design distribution and the infill criteria used in the adaptive steps of the SMBO framework. They compare performances on two variants of the Branin function, a classic benchmark in SMBO research, and on two parameter tuning problems. They conclude that the total budget has an important influence on the ranking of the different SMBO algorithms. In line with our observations and conclusions, they recommend tuning of the SMBO design if one is likely to see similar types of problems several times. 

More recently, Lindauer et al.~\cite{Lindauer19} analyze the sensitivity of Bayesian optimization heuristics w.r.t.~its own hyper-parameters. This study, however, puts a much stronger emphasis on the various design choices, and details for the initial sampling strategy are not explicitly mentioned, although Table~3 in their work suggests that this has been varied as well.

\vspace*{-0.025cm}
\paragraph{Availability of Project Data} 
While this report highlights a few of our key findings, and demonstrates which statistics are possible to obtain with the data, the full data base offers much more than we can touch upon in a single conference paper. Not only can our data be used to zoom further into the various settings described below, but it also offers additional information about 
%the discrepancy of the initial point sets, 
the function value of the best initial design point and of the first point queried in an adaptive fashion, as well as the distance of these points and of the best solution to the optimal solution (in the decision space $[-5,5]^d$, measured in terms of the L2 norm).

Please note that most of the results reported below are based on median values per (dimension, function, total budget, initial budget ratio, design) combination. This is to avoid correcting factors for the comparison between the Halton designs (for which we have 5 runs for each of the $7\,200$ considered settings) and the other three designs (for which we have 25 independent runs per setting, i.e., 5 SMBO runs for each of the 5 random samples from the design). Detailed results for each experiment are available in the data base, so that one can easily perform statistical tests, or use other aggregation methods. An interactive evaluation of the data is possible with the very recently released tool HiPlot~\cite{HiPlot}, which essentially produces parallel coordinate plots through which one can easily navigate by zooming and/or highlighting different parts of the data. 

The interested reader can find all our project data on~\cite{data}.

\section{Sequential Model-Based Optimization}
\label{sec:smbo}

In many real-world applications like production engineering, numerical simulations, or hyper-parameter tuning, the objective function $f$ at hand is often of black-box nature. That is, (a) there is little or no knowledge about the structure of $f$ (in particular, we typically do not have derivatives), and (b) function evaluations are expensive in terms of computational and/or monetary resources (days of computation time or actual physical experiments). As a consequence, in the course of problem solving, one tries to keep the number of true function evaluations low. In such settings, sequential model-based optimization (SMBO, \cite{HWBWB2015}) -- also known under the term Bayesian optimization\footnote{Originally, Bayesian Optimization only referred to SMBO approaches with Bayesian priors, but nowadays the term is often used to denote the whole class of SMBO methods.} -- advanced to the state-of-the-art in recent years and is used extensively in many fields of research, e.g., within versatile tools for automatic algorithm configuration~\cite{SMAC}.

In a nutshell, the key idea of SMBO is as follows: a regression model, i.e., an approximation $\hat{f}$ to the true optimization problem $f$, is fitted to the evaluated points of an initial design. Subsequently, the model $\hat{f}$ serves as a cheap surrogate for the expensive true objective function and is used to determine the next point(s) worth being evaluated through the actual problem $f$. These points are determined by optimizing a so-called \emph{infill criterion} (also referred to as \emph{acquisition function}) which keeps balance between exploiting the model (in the sense of striving to high-quality points) and exploring the search-space regions which lack a good model fit (i.e., regions with a high uncertainty about the quality of approximation $\hat{f}$). Note that the optimization of the acquisition function itself is an (often highly multimodal) optimization problem, which is typically solved by state-of-the art solvers such as CMA-ES~\cite{HansenO01}, Nelder-Mead~\cite{NelderM65}, or simply by standard Newton methods, if the surrogate model $\hat{f}$ allows. The key here is that those algorithms now operate on $\hat{f}$ and not on $f$, which can be evaluated much more efficiently. The points proposed from the optimization of the acquisition function are then evaluated through $f$ and the surrogate $\hat{f}$ is updated to account for the new information.  
The process is repeated until the available budget of time or function evaluations is depleted. 

Jones~\cite{JonesSW98} was the first who used this approach in his Efficient Global Optimization (EGO) algorithm. Therein, Gaussian processes serve as the surrogate and expected improvement (EI) is adopted as infill criterion. Following Jones' seminal contribution, a plethora of extensions were proposed by the community including multi-point proposal~\cite{Bischl2014MOIMBO} and multi-objective SMBO (e.g.,~\cite{Knowles2006ParEGO}) making SMBO a highly flexible framework with many interchangeable components and facets. We refer the interested reader to~\cite{HWBWB2015} (and references therein) for a comprehensive overview. 

Our study is based on the classical EGO algorithm by Jones.

\section{Experimental Setup}
\label{sec:expsetup}

Our study investigates the effect of the total budget, the size of the initial design (i.e., the number of evaluations prior to building the first surrogate), and the distribution of this initial design on the quality of the final recommendation made by an off-the-shelf SMBO algorithm. 
Below, we summarize the benchmark problems and solution strategies (Section~\ref{sec:benchmark}), as well as the performance measures that we used to evaluate the different strategies (Section~\ref{sec:performance}). 

All our experiments are implemented in the R programming environment~\cite{Rcore2018}. To be more precise: the SMBO framework \RmlrMBO~\cite{BRB2017mlrMBO} serves as the working horse for our experimental study, the \Rsmoof-package~\cite{Bossek2017Smoof} is used for an interface to the BBOB functions and the interface package \Rdandy~\cite{Bossek2020dandy} is used to generate the initial designs. 
The latter delegates to packages \Rqrng~\cite{Rqrng} and \Rrandtoolbox~\cite{Rrndtoolbox}, which implement quasi-random sequence generators as well as to package \Rlhs~\cite{Rlhs} for the LHS designs.

\subsection{Benchmark Problems and Solvers}
\label{sec:benchmark}
We use the following setup for our experimental analysis: 
\begin{itemize}[leftmargin=7pt]
    \item \textbf{The objective function $f$.} As mentioned in the introduction, we focus on the 24 functions from the (noiseless and single-objective) BBOB test suite~\cite{hansen2016cocoplat}. An overview of these functions is available in~\cite{Hansen2009_Noiseless}. We consider the first instance of each function, whose $d$-dimensional variant we denote by $f^{(d)}$. We let $F^{d}$ be the collection of these 24 functions. We study \emph{minimization} as objective.  
    \item \textbf{The problem dimension $d$.} We consider five different search space dimensions: $d \in D := \{2, 3, 4, 5, 10\}$.
    \item \textbf{Total budget $n$.} The total number of function evaluations. We consider six different budgets: $n \in N := \{2^x \mid x \in \{4, \ldots, 9\}\}$.
    \item \textbf{Initial design ratio $k$:} We consider initial designs of size $\lceil k\cdot n \rceil$ with $k \in K:=\{0.1, 0.2, \ldots, 1.0\}$.
    \item \textbf{Sampling design $s$.}  
    We study four different distributions from which the $d$-dimensional initial design of size $\lceil k\cdot n \rceil$ is sampled: 
    \begin{itemize}
        \item \emph{uniform sampling}: R's default random number generator (Mersenne-Twister~\cite{MersenneTwister}) to generate uniform samples. 
        \item \emph{Latin Hypercube Sampling} (LHS~\cite{LHS}): ``improved'' LHS design as suggested in~\cite{beachkofski_improved_nodate}. %(\texttt{lhs::improvedLHS)})
        \item \emph{Sobol' sequences}~\cite{Sobol}: \Rrandtoolbox implementation with scrambling as proposed by Owen~\cite{Owen1995}, and Faure \& Tezuka~\cite{FaureT2002}.
        \item \emph{Halton designs}~\cite{Halton60} \Rrandtoolbox implementation with default parameters.
    \end{itemize}
    More detailed definitions, motivations, and applications of these distributions can be found, for example, in~\cite{DickP10}.

    \item \textbf{Random seed $r_i$ - initial design.} While the Halton point sets are deterministic, the other designs produce random points. To account for this randomness, we sample $R_i=5$ instances from each of the three random (i.e., non-Halton) designs.   
    \item \textbf{Random seed $r_A$ - SMBO randomness.} Finally, to compensate for the randomness of the SMBO algorithm (note that the SMBO process is stochastic itself, e.g., by means of a stochastic procedure used to search the infill-criterion), we do $R_A=5$ independent runs per each of the settings fixed through the decisions above.
\end{itemize}
    
It should be noted that we do not vary the infill criterion (also known as \textit{acquisition function}), nor any other component of the SMBO, but use the default variant of \RmlrMBO v1.1.4 with expected improvement as infill criterion and a Kriging surrogate.

With the notation above, we consider a total number of $|F| \cdot |D| \cdot |N|=24\cdot 5 \cdot 6 =720$ different \textbf{problems}, and for each of these problems we consider $|S| \cdot |K| = 4 \cdot 10 = 40$ different solution \textbf{strategies}. Here we consider the budget as integral part of a problem, since SMBO algorithms are typically applied when the budget is fixed a priori. We therefore distinguish between the \emph{function} $f^{(d)}$ that is to be optimized, and the \emph{problem} $(f,d,n)$ of minimizing $f^{(d)}$ with a given budget $n$. 

As mentioned above, on each problem we perform 5 runs of each strategy which is based on Halton designs and we perform 25 runs for all other strategies. 
Our total number of experiments is thus 
\begin{align*}
  & |F| \cdot |D| \cdot |N| \cdot |K|  \cdot (1+(|S|-1)\cdot R_i)) \cdot R_A \\
  = 
  &  24 \cdot 5 \cdot 6  \cdot 10 \cdot (1+(3 \cdot5))  \cdot 5 
  = \underline{576\,000}.  
\end{align*}
Not all of these runs terminated successfully, due to problems with the Kriging implementation used by \RmlrMBO. The problems occur in particular with high total budget and low initial design ratio. Here, the Kriging-routine obviously runs into problems when many points are sampled close to each other as it often is the case when SMBO runs converge into a (local) optimum. While for each $n \leq 128$ there are at least $99.8\%$ successful runs this number reduces to $94\%$ for $n=256$ and $85.4\%$ for $n=512$.
In total, we had $555\,598$ $(96.5\%)$ successful runs. In all computations below we only consider $(f,d,n,k,s)$ combinations for which at least three runs terminated successfully, i.e., provided their recommendation.

\subsection{Performance Measures and VBS}
\label{sec:performance}

For each of our experiments we record the value of the best solution that has been evaluated during the entire run. We denote this value by $f(d,n,k,s,r_i,r_A)$. Since the BBOB functions have quite diverse ranges of function values, we do not study these function values directly, but rather follow standard practice in BBOB studies and focus on the \emph{target precision}, i.e., the gap to the global optimum, 
$$p(f,d,n,k,s,r_i,r_A):=f(d,n,k,s,r_i,r_A) - \inf f^{(d)}.$$

As mentioned above, we will restrict most of our analyses to the median performance of each strategy on each problem. Our main performance criteria is therefore 
\begin{align*}
    % & M(f,d,n,k,s,r_A)  = \M\left( \{ p(f,d,n,k,s,r_i,r_A) \mid r_i \in R_i(s) \} \right), \text{ and}\\ %brauchen wir aktuell nicht 
    & M(f,d,n,k,s)  = \M\left( \{ p(f,d,n,k,s,r_i,r_A) \mid r_i \in R_i(s), r_A \in [5] \}\right),
\end{align*}
where $\M$ denotes the median and where we use the convention that $R_i(\text{Halton})=\{1\}$ and $R_i(s)=\{1, 2, \ldots, 5\}=:[5]$ for the other sampling designs $s \in S\setminus\{\text{Halton}\}$. 

\paragraph{Virtual Best Solver and Relative Target Precision}
% \label{sec:VBS}

\begin{figure}[t]
    \centering
    \includegraphics[width=0.9\columnwidth, trim=0 7pt 0 11pt, clip]{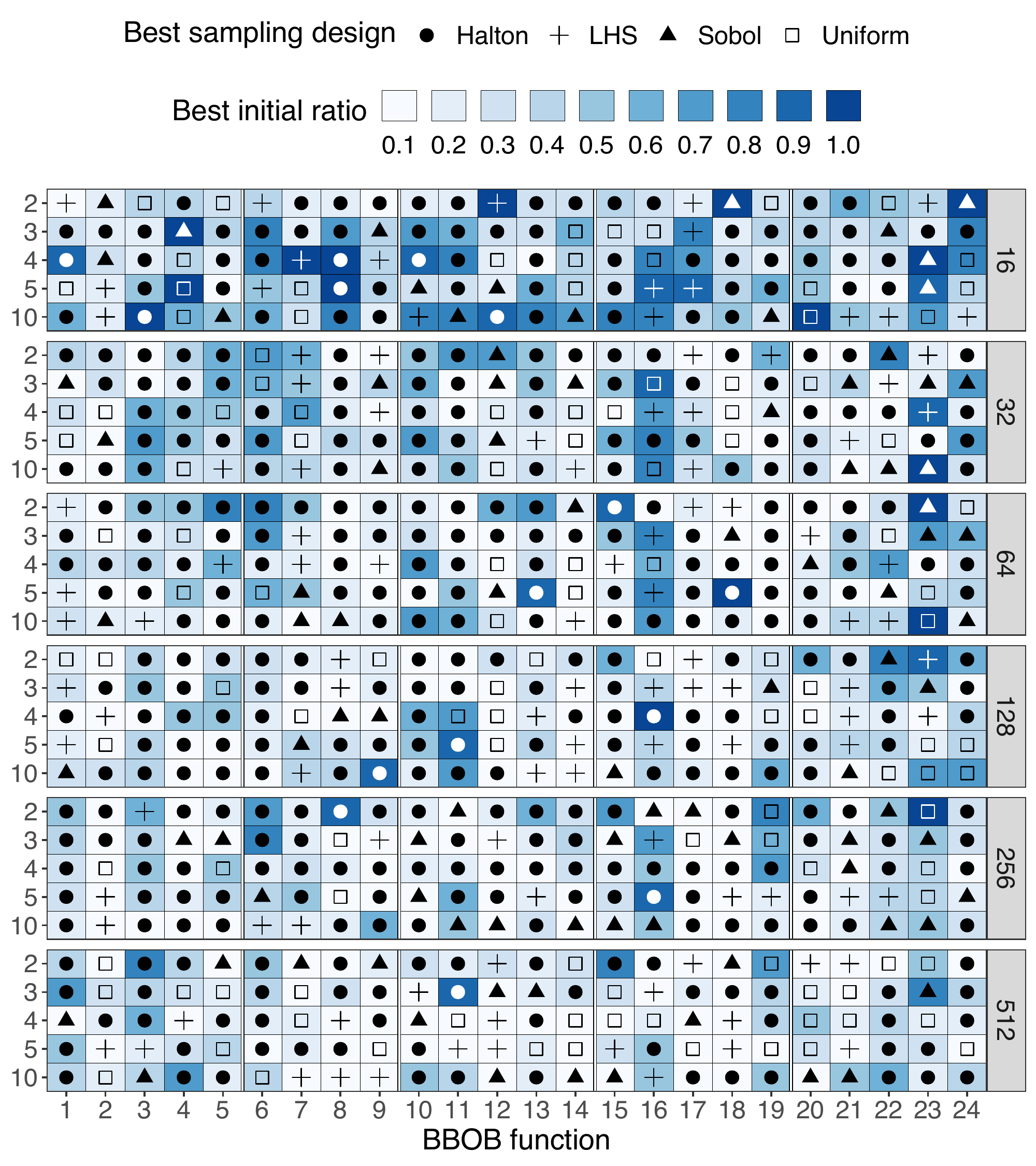}\vspace*{-0.15cm}
    \caption{Overview of the virtual best solver (VBS), i.e., the strategy $(k,s)$ that achieved the best median performance on the respective problem $(f,d,n)$.
    }\vspace*{-0.15cm}
    \label{fig:vbs-per-dim}
\end{figure}

An important concept in comparing portfolios of algorithms is the \emph{virtual best solver} (VBS). This VBS describes a hypothetical algorithm that for each problem (i.e., each $(f,d,n)$ combination in our case) selects an algorithm $A$ from a given portfolio $\mathcal{A}$ that achieves the best performance \cite{kerschke2018survey}. In our case, the algorithm portfolio is the collection of all 40 $(k,s)$ combinations. As we consider median performance, the VBS is defined by selecting for each problem $(f,d,n)$ the strategy $(k,s)$ that achieved the best median function value. For notational convenience, we omit the explicit mention of the median and set 
\begin{align*}
    \VBS(f,d,n):= \min \{M(f,d,n,k,s) \mid s \in S, k \in K\}.
\end{align*} 
Fig.~\ref{fig:vbs-per-dim} shows which strategy defined the VBS for which problem(s). A first visual interpretation suggests that this data is relatively unstructured; we will come back to this point further below. 

By design, some of the BBOB functions are much ``harder'' than others, so that we see substantial differences in the target precision that can be achieved with a fixed budget $n$. To compensate for that in our aggregations, we will frequently study the relative performance of a strategy $(k,s)$ compared to the VBS. To this end, we set 
\begin{align*}
    \Er(f,d,n,k,s):= M(f,d,n,k,s)/\VBS(f,d,n) 
\end{align*}
and refer to $\Er(f,d,n,k,s)$ as the \emph{relative target precision} of strategy $(k,s)$ on problem $(f,d,n)$. Note that these values are at least one, where $\Er(f,d,n,k,s)=1$ implies that strategy $(k,s)$ achieved the best median target precision among all the 40 different strategies. 

\section{Aggregated Results}
\label{sec:results-overall}

As shown in Fig.~\ref{fig:vbs-per-dim}, it is not possible to derive simple rules that define which strategy achieves the best performance on each of the BBOB functions. In Fig.~\ref{fig:simply-the-best} we therefore count how often each strategy forms the VBS. Therein, we observe a clear advantage for Halton designs (it has the most ``hits'' for any given initial ratio except for $k=100\%$), and we further observe a clear tendency towards small initial ratios. However, we also see that each strategy ``wins'' at least one problem. 
Neither the simple counting statistics in Fig.~\ref{fig:simply-the-best} nor the more detailed overview in Fig.~\ref{fig:vbs-per-dim} provide any information about the \emph{magnitude} of the advantage. We thus plot in Fig.~\ref{fig:overall-boxplots} the distribution of the relative target precision $\Er(f,d,n,k,s)$ of each strategy $(k,s)$, aggregated again over all 720 problems. This plot confirms the tendency that spending a larger ratio of the total budget on generating the initial design results in worse overall performance. We also observe that although the \textbf{Halton design generated with $\mathbf{k = 10\%}$ of the total budget has the best median performance}, the actual differences between the four designs are rather small.

\begin{figure}[t]
    \centering
    \includegraphics[width=\linewidth, trim=0 5pt 0 11pt, clip]{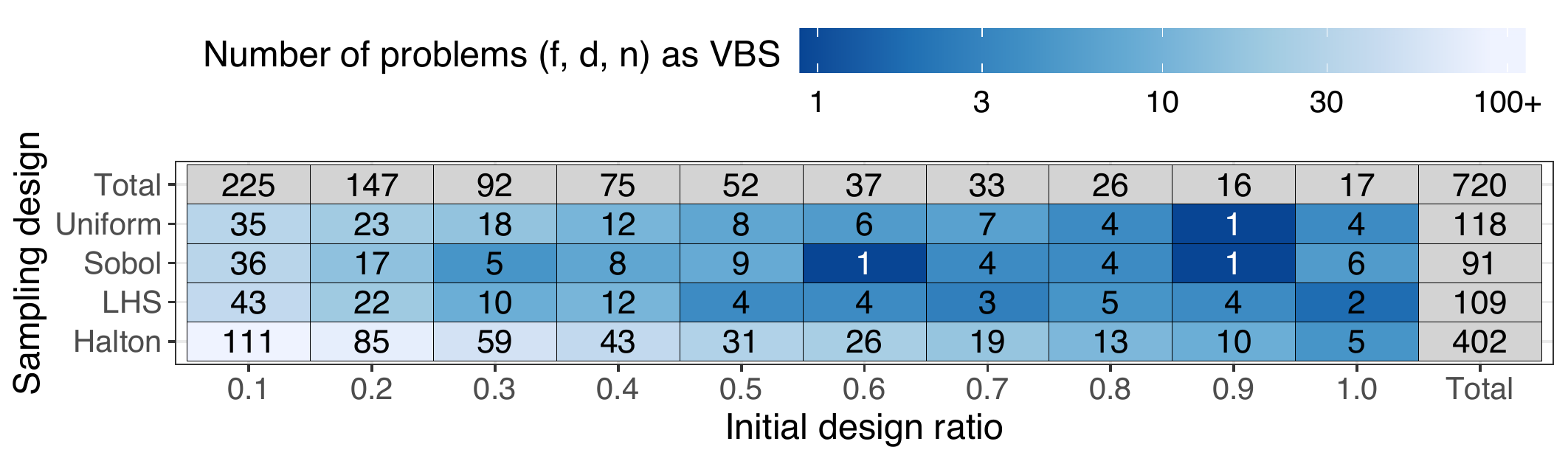}\vspace*{-0.2cm}
    \caption{Number of problems $(f,d,n)$ for which the respective strategy forms the VBS. 
    }\vspace*{0.25cm}
    \label{fig:simply-the-best}

    \includegraphics[width=\linewidth, trim=0 5pt 0 11pt, clip]{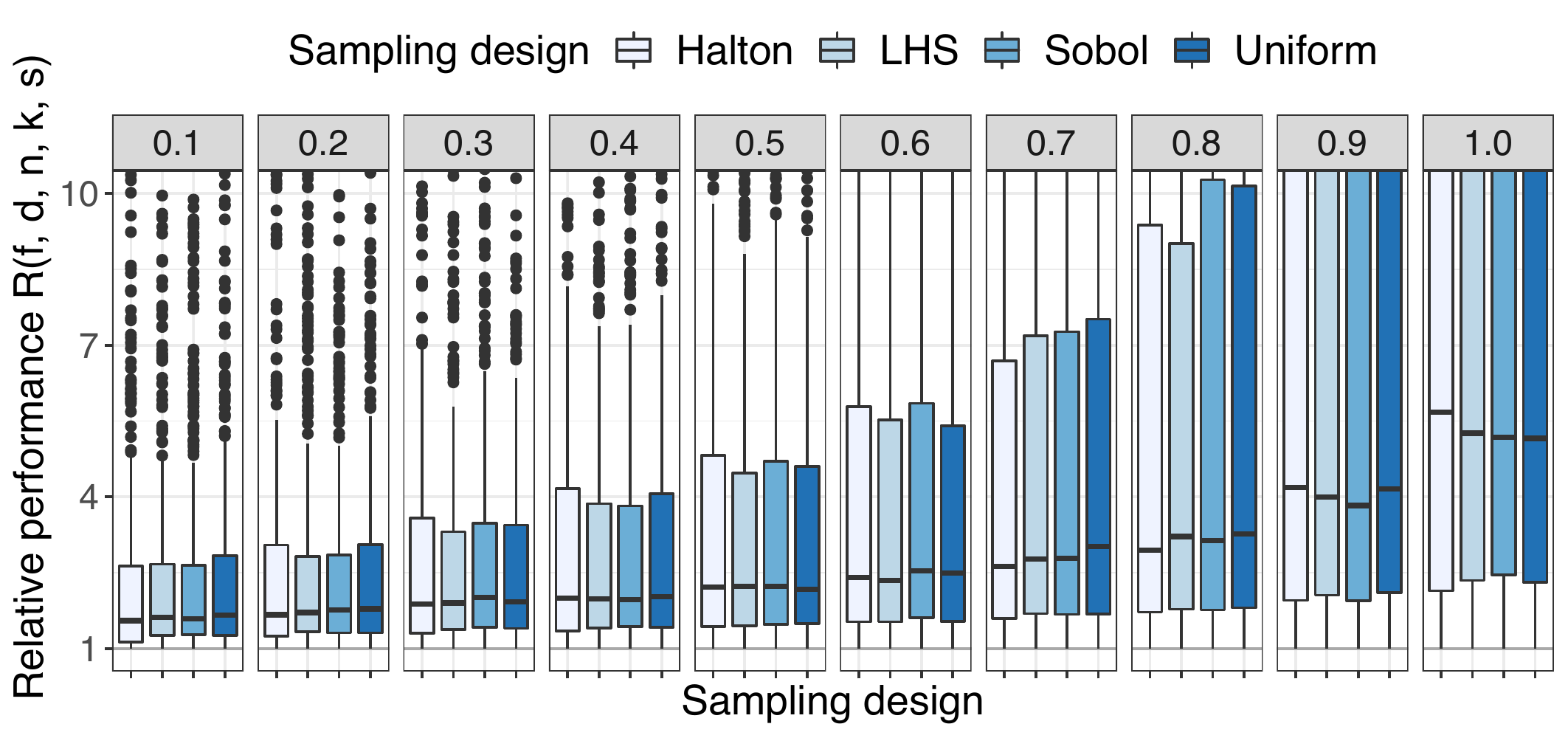}\vspace*{-0.2cm}
    \caption{Boxplots of relative performances $\Er(f,d,n,k,s)$ across all 600 problems $(f,d,n)$ with budget $n \le 256$, shown for all 40 different strategies $(k,s)$. The $y$-axis is capped at 10. 
    }\vspace*{-0.15cm}
    \label{fig:overall-boxplots}
\end{figure}

\begin{figure}[ht]
    \centering
    \includegraphics[width=0.93\linewidth, trim=0 5pt 0 11pt, clip]{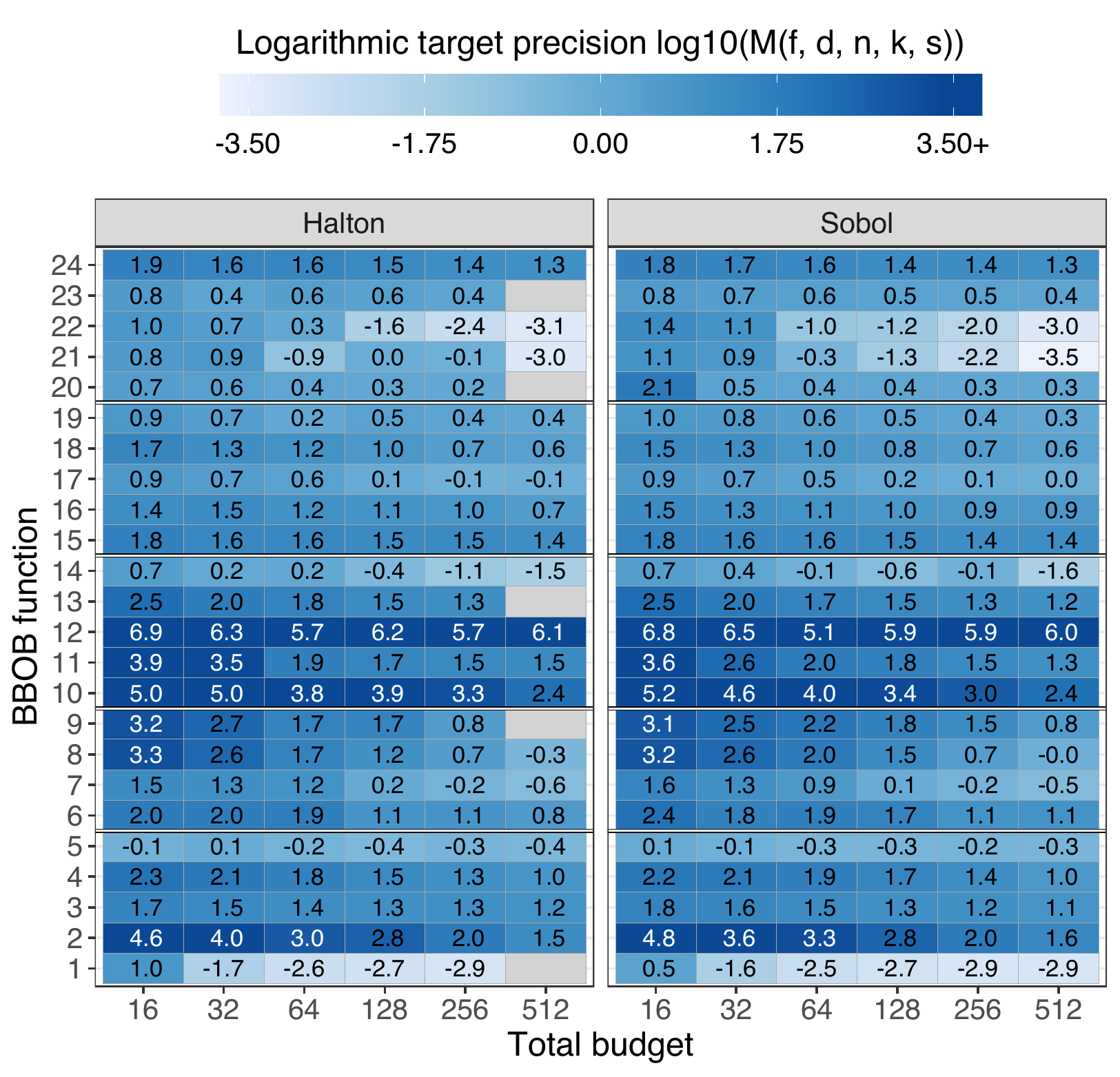}
    \caption{Logarithmic median target precision $\log_{10}(M(f,d,n,k,s))$ depending on the total budget. Results are shown for Halton (left) and Sobol (right) designs with an initial budget of 10\% of the total budget and across all 5-dimensional BBOB functions. Gray boxes are due to missing data (less than 3 successful runs, see Section~\ref{sec:benchmark}).}
    \label{fig:convergence}
\end{figure}

\begin{figure*}[ht]
    \centering
    \includegraphics[width=\linewidth, trim = 2mm 1mm 2mm 1mm, clip]{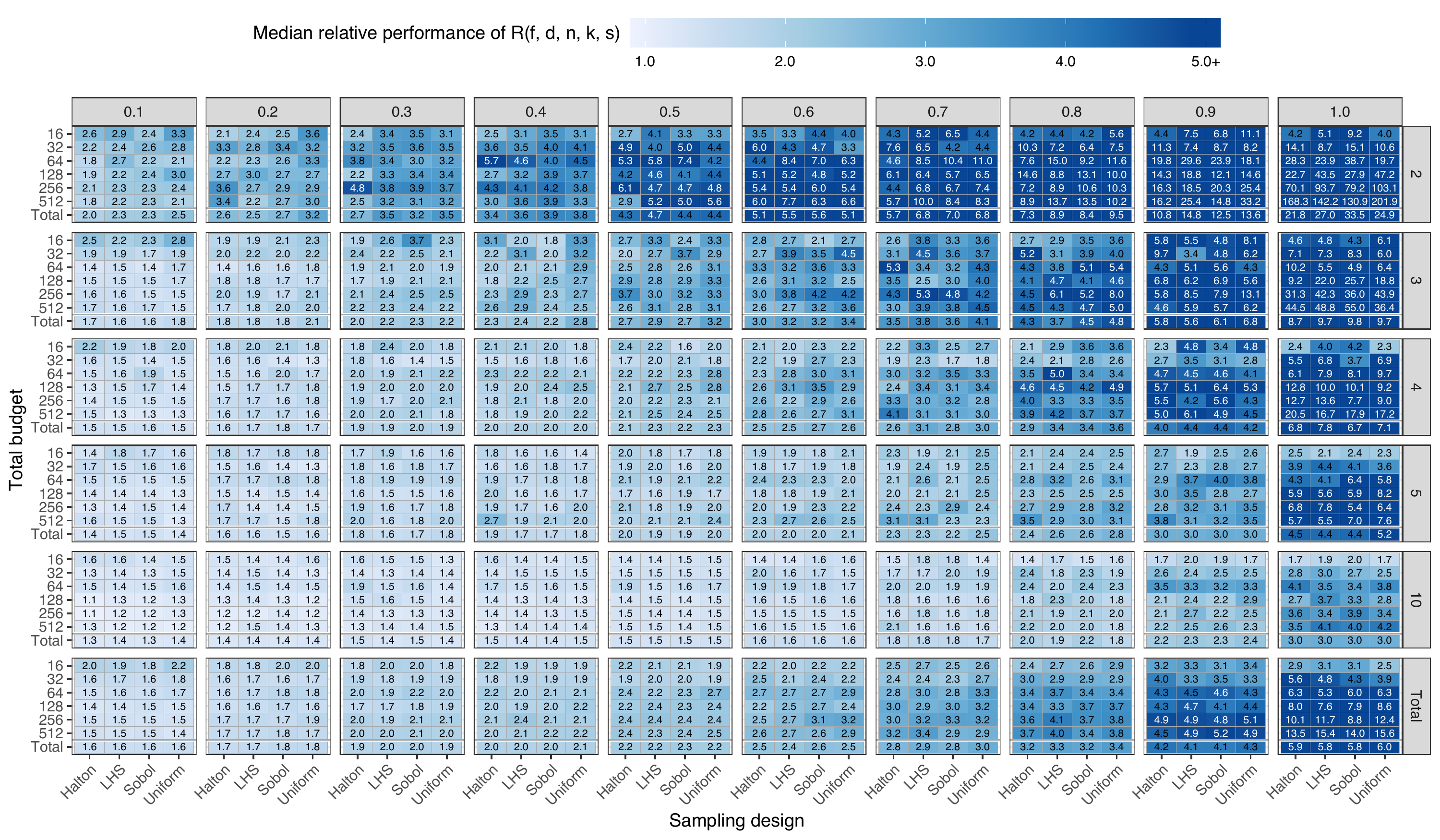}
    \caption{Median (over all 24 BBOB functions) relative performance of $\Er(f,d,n,k,s)$, by dimension and budget (rows) and strategy $(k,s)$ (columns). 
    }
    \label{fig:overall-relative}
\end{figure*}

A more detailed picture about the relative performances is provided in Fig.~\ref{fig:overall-relative}. Here, we plot the median (over all 24 BBOB functions) relative performance; i.e., the value in each cell represents $\M\left(\{ \Er(f,d,n,k,s) \mid f \in \{1, 2, ..., 24\} \}\right)$ for the given dimension, budget, and strategy. 
We observe that in most cases the \textbf{performance worsens with increasing initial budget ratio $\mathbf{k}$}, and this consistently for each problem dimension $d$ and total budget $n$. 

The values in the rows labeled ``Total'' are the median values over all budgets (last row per dimension) and dimensions (bottom-most rows), respectively. Noticeably, the \textbf{influence of the sampling design vanishes with increasing dimension} -- independent from the budget ratio. Aggregated over all dimensions, the differences between the designs are small, as already observed in Fig.~\ref{fig:overall-boxplots}.

Remember that the values in Fig.~\ref{fig:overall-relative} are always scaled by the VBS that is specific for problem $(f,d,n)$, but independent from strategy $(k,s)$. This implies that the rows are computed against the same VBS, but different rows compare against different strategies. Values in different rows should therefore only be compared with care.

%%%%%%%%%%%%%5
\section{Performance by Function}
\label{sec:results-details}

After having studied values that were aggregated across all 24 BBOB functions (see Section~\ref{sec:results-overall}), we now take a closer look at the differences between the different strategies on each of the functions.

\paragraph{Influence of the Total Budget} Fig.~\ref{fig:convergence} reports the median target precision (shown on a log-scale) achieved by Halton and Sobol' designs with $k = 10\%$ initial budget, in dependence of function $f$ and total budget $n$. The plot reveals the functions that are easy (e.g., functions 1, 21, 22) and difficult (functions 10 and 12) for SMBO. 
Note that the performance convergence is not always monotonically decreasing with increasing total budget size. This might result from the small number of repetitions (5 for the Halton design, 25 for Sobol'). However, the differences are fairly small. 
Fig.~\ref{fig:heatmap_approx_error_min_median_by_method_and_totalbudget} extends Fig.~\ref{fig:convergence} to all 40 strategies $(k, s)$. That is, for each 5-dimensional problem $(f,5,n)$ a heatmap of the relative performances $\Er(f,d,n,k,s)$ is shown for all pairs of sampling design $s$ and initial design ratio $k$. We observe that, in particular for functions 15, 19, 23 and 24, the differences between the different initial budgets are comparatively small. This likely results from the functions' highly multimodal landscapes, which hinder SMBO from training reasonable surrogates.

\begin{figure*}[htbp]
    \centering
        \includegraphics[width=\textwidth, trim=0 5pt 0 11pt, clip]{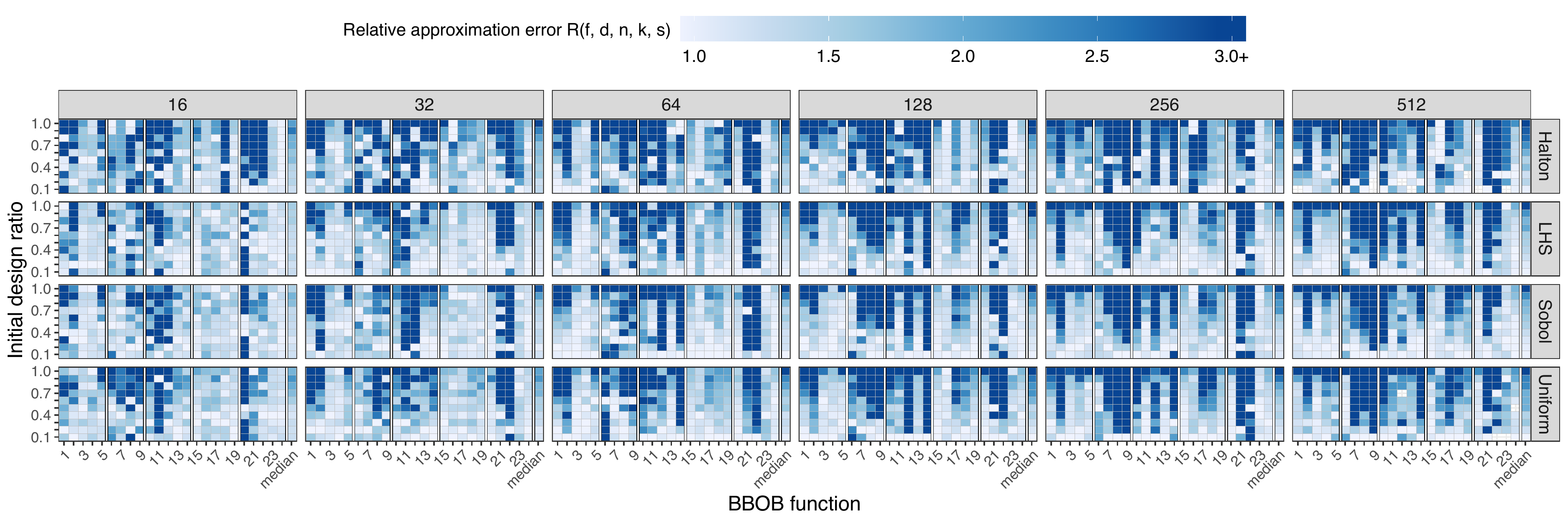}
    \caption{Heatmap visualization of relative performances $\Er(f,d,n,k,s)$ by function, total budget, and strategy $(k,s)$ for fixed dimension $d=5$. Values are capped at 3.}
    \label{fig:heatmap_approx_error_min_median_by_method_and_totalbudget}

    \includegraphics[width=\textwidth, trim=0 5pt 0 11pt, clip]{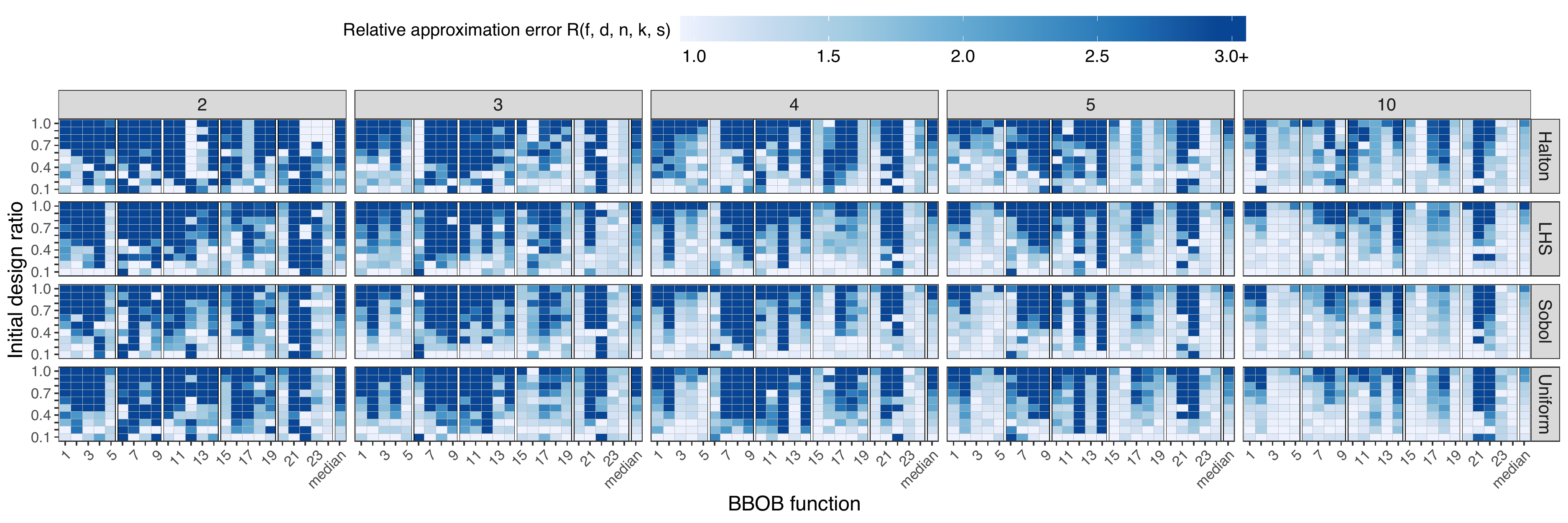}
    \caption{Heatmap visualization of the relative performance $\Er(f,d,n,k,s)$ by dimension, function, and design type for a fixed total budget of $n=128$ function evaluations. Values are capped at 3.
    }
    \label{fig:heatmap_approx_error_min_median_by_idr_and_method}
\end{figure*}

\paragraph{Influence of the initial sample size $k$ and design $s$} 
Fig.~\ref{fig:heatmap_approx_error_min_median_by_idr_and_method} shows the relative median target precision $\Er(f,d,n,k,s)$ for all 24 BBOB functions, for a fixed budget of 128 function evaluations and variable dimension (columns) and strategies (rows). We recall that the VBS is defined per column, i.e., each column has at least one strategy with $\Er(f,d,n,k,s)=1$ (see Fig.~\ref{fig:vbs-per-dim}).  

We observe that the benefit of small initial budgets is important for functions with at most medium-sized indices. This finding is very plausible, as the first 14 functions mainly are separable and/or unimodal -- i.e., functions whose structure can be well exploited by SMBO. However, \textbf{for the group of multimodal functions} (IDs 15 to 24), with the notable exception of functions 21 and 22, the differences between the different initial ratios are rather small, indicating that \textbf{SMBO does not perform much better than \mbox{(quasi-)} random sampling} in the initial phases of the optimization process. 

We also see interesting cases in which larger ratios of initial budget result even in better performance than small initial ratios. An extreme case is function 12 in dimension $d=2$. Its situation is as follows. The VBS is defined by the (30\%, Halton) strategy. The differences between the Halton designs with $k>30\%$ are rather small, whereas for the other strategies smaller initial budgets are preferable. By studying the absolute values in more detail, we find that the Halton strategy identifies a point with absolute target precision $24.9$ when $k\ge 30\%$. SMBO does not manage to find a better point in any of its $128-\lceil 30\% \cdot 128 \rceil = 89$ adaptive evaluations. The best median target precision of any of the other strategies has target precision $58.6$ -- achieved by the (10\%, LHS) strategy. Looking further into the results of the 800 individual runs, we find that 126 of them find a point of target precision smaller than $24.9$. The distribution of their initial ratios is not unanimous, as can be see in the following table, which counts how often each initial ratio $k$ appears among these 126 runs. These results show how difficult it is to give a general advice for the optimization of this function -- even when the budget is fixed and the function ID known. %\\
\begin{center}
  \begin{tabular}{l|cccccccccc}
 $k$& 0.1 & 0.2 & 0.3 & 0.4 & 0.5 & 0.6 & 0.7 & 0.8 & 0.9 & 1  \\
% $k$& 10\% & 20\% & 30\% & 40\% & 50\% & 60\% & 70\% & 80\% & 90\% & 100\%\\
\hline
\# & 14  & 12  & 8   & 13  & 21  & 7   & 12  & 14  & 10  & 15
\end{tabular}  
\end{center}
% \linebreak \noindent

\section{Restarts vs. Long Runs}
\label{sec:restart}

In the previous paragraph, we have started to look into the distribution of the target precisions. We now demonstrate how such  information can be used to study whether it is beneficial to use the total budget of $n$ function evaluations for a single long run, or whether one should rather start two shorter runs of budget $n/2$ each, or four runs of budget $n/4$, etc. 

\paragraph{Distribution of the Target Precisions}
% \label{sec:distribution}

\begin{figure}
    \centering
    \includegraphics[width=\linewidth, trim=0 5pt 0 11pt, clip]{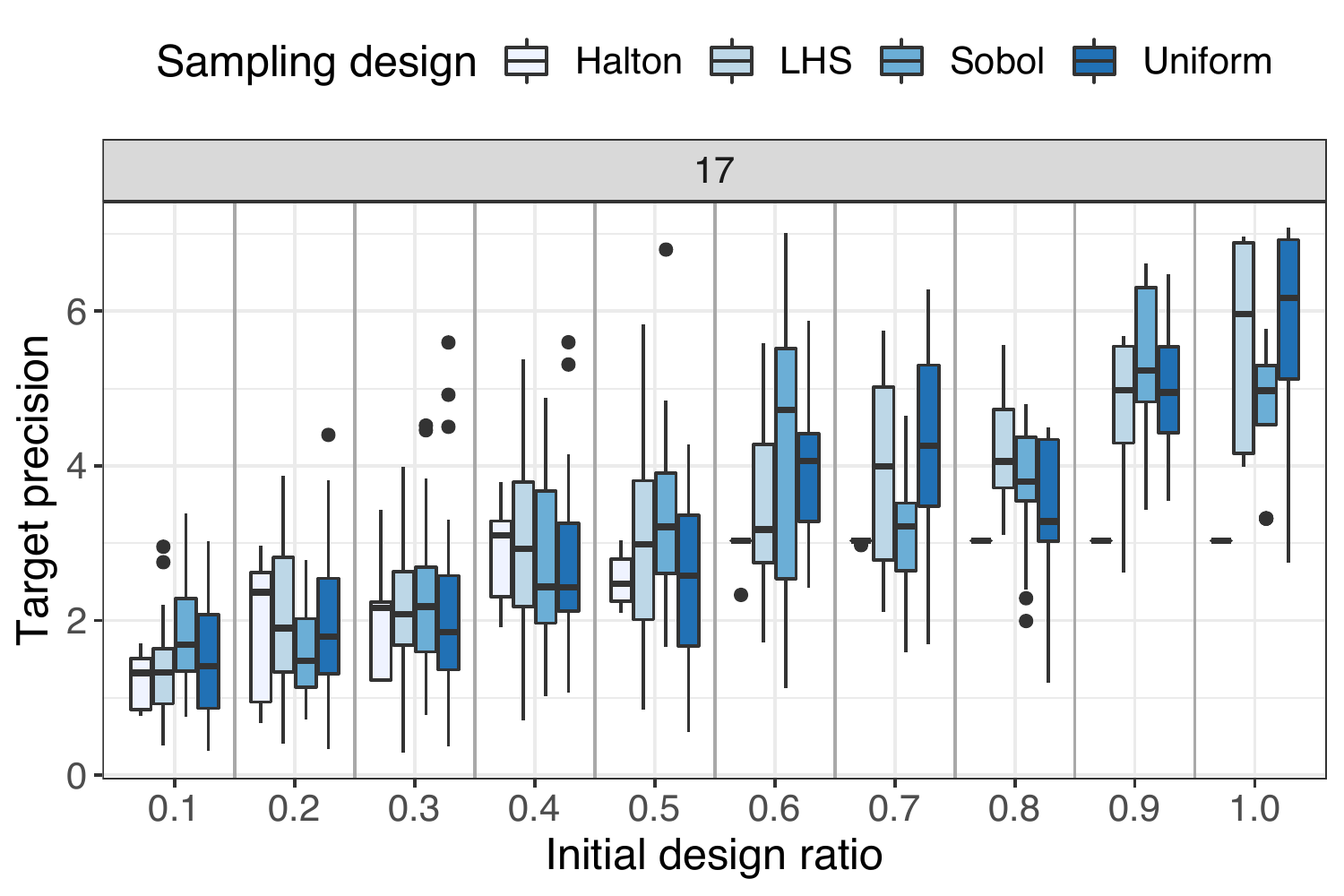}
    \caption{Boxplots of the target precisions $p(17,5,128,k,s,r_i,r_A)$-values for function $f=17$, dimension $5$, and total budget $n=128$, grouped by initial budget ratio $k$ and design $s$. 
    }
    \label{fig:f17-boxplot}
\end{figure}

Crucial for the consideration of restarts are the distributions of the function values (or, equivalently, the distributions of the target precisions) achieved  by the different strategies $(k,s)$. For reasons of space, we cannot go in much detail here, but Fig.~\ref{fig:f17-boxplot} demonstrates how these boxplots look like. Note that this figure is for one specific combination of function ($f = 17$), dimension ($d=5$) and budget ($n=128$). It aggregates the target precision of all 40 strategies, i.e., of 800 runs in total. Our data base contains one such plot for each of the 720 $(f,d,n)$ problems.

Note that the dispersion of Halton designs are smaller, but this is due to the fact that we do not perform resampling for this sequence. For all pairs of ($k$,Halton) strategies with $k\ge 50\%$ the target precision of the best initial design point is slightly above 3. For $k\ge 80\%$ none of the SMBO runs starting in this best initial design point finds a solution of better target precision. For $k \in \{60\%, 70\%\}$, only one of the five runs each finds a better solution. Note that the length of each of these SMBO runs is $n-\lceil k \cdot n \rceil$, which for $k=0.6$ corresponds to 51 adaptive SMBO steps. Such detailed information could be very \textbf{useful to identify weaknesses of the EGO approach}, and, hopefully, contribute towards better SMBO designs.

\paragraph{Computing median target precision of restarting SMBO} To investigate if, for a given problem $(f,d,n)$, a restart strategy is beneficial over a single long run, we need to extend our previous focus on median target precision to different percentiles. To this end, let
 \begin{align*}
     P_q(f,d,n,k,s) := P_q \left( \{p(f,d,n,k,s,r_i,r_A) \mid r_i \in R_i(s), r_A \in [5]\}\right),
 \end{align*}
 the $q$-th percentile of the target precisions achieved by strategy $(k,s)$ on problem $(f,d,n)$ across all 5 (Halton) or 25 (Sobol', LHS, uniform designs) runs, respectively. For a fair comparison of one run of the full budget $n$ with two runs of budget $n/2$ (of the same strategy), we compare the median $M(f,d,n,k,s)$ (i.e., the 50-th percentile) with the $q:=1-\sqrt{1/2}$-th percentile $P_q(f,d,n/2,k,s)$. With this value of $q$, the probability that (at least) one of the two runs achieves a target precision that is at least as good as $P_q(f,d,n/2,k,s)$ equals $1-(1-q)^2=1/2$. This is identical to the probability that one long run achieves a target precision that is at least as good as $M(f,d,n,k,s)$. Note that we disregard a small bias in our data, which results from the fact that we do not have $25$ completely independent runs. Instead, we use the same initial design sample for five independent SMBO runs each -- but, we ignore this effect in the following computations. Also, given the small number of runs, all numbers should be taken with care -- the smaller the percentile, the larger the uncertainty around the values. We nevertheless show this example to demonstrate how one could systematically address the question how to split a given budget into possibly parallel runs.

 Fig.~\ref{fig:heatmap_percentiles_2d} illustrates an example for the relevant percentiles when comparing one long run of budget $n$ with two short ones of budget $n/2$, and four even shorter ones of budget $n/4$. More precisely, we fix in this figure the strategy to (10\%,LHS) and the dimension to $d=5$, and we show log-scaled relative data. Each $3\times 6$ box corresponds to one of the 24 BBOB functions. As we scale the values within a box by its VBS, and afterwards show the percentile ratios on a log-scale, the field with value 0.0 represents the combination achieving the best target precision (i.e., the VBS) among the displayed combinations. Not surprisingly, for most functions this is the $(1-\sqrt[4]{1/2})$-th percentile of the full budget $n=512$. Let $P_f^*$ be the target precision of this (percentile, budget) combination for a given function $f$. A value $\varphi$ in field $(q,n)$ is then to be read as follows: the target precision $P_q(n):=P_q(f,d=5,n,10\%,LHS)$ satisfies $P_q(n)=10^{\varphi} \cdot P^*_f$. Smaller values are therefore better. We see that for $f = 5$,% F1$^{(5)}$, 
 for example, our data suggests that a total budget of 512 evaluations (value 0.4 when used as single run) is better used for four runs of budget 128 each (value 0.1). We have marked in this matrix all fields for which the long run compares unfavorably with a restart strategy -- the one corresponding to the neighboring field on the lower left diagonal. Overall, we see that several such cases exist, which confirms our previous finding that \textbf{EGO does not always compare favorably against quasi-random sampling}.

\begin{figure}
     \centering
    \includegraphics[width=\linewidth, trim=0 5pt 0 11pt, clip]{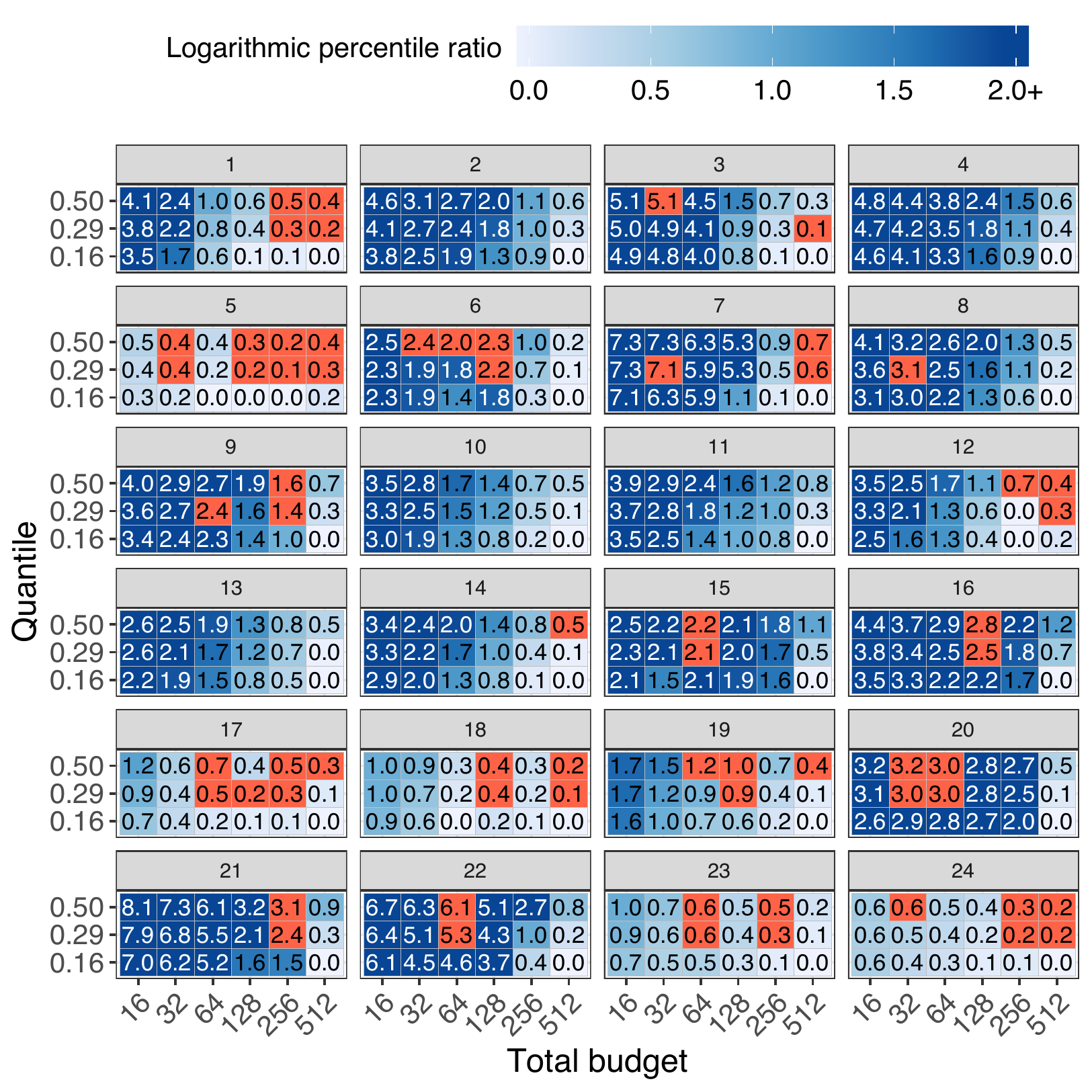}
     \caption{Percentiles $P_q(f,d,n,k,s)$ of target precisions across the 25 SMBO runs per function and dimension using an LHS design with 10\% initial budget and for the 2-dimensional problems. The percentiles are scaled by the respective function's best percentile, and the resulting ratios are shown on a capped log10-scale. Red boxes indicate that the corresponding strategy performs unfavorably against a restart strategy (the one to the lower left).  
     }
     \label{fig:heatmap_percentiles_2d}
 \end{figure}
 
\section{Conclusions}
In this paper we have presented a database for data-driven investigations of the sequential model-based optimization (SMBO) strategy EGO~\cite{JonesSW98}. The focus of our work is on analyzing the influence of the (size and type of) initial design on the overall performance of EGO. Our data base contains data for 720 different problems, which are evaluated against a total of 40 different initial design strategies. 

While we clearly observed that small initial designs are preferable at a high-level view, we also found that each of the 40 considered combinations of design type and size achieved best performance on at least one of the 720 problems. Our findings thus confirm that an automated strategy selection method -- like the proof-of-concept approach presented in \cite{saini2019automatic} -- might indeed be profitable. Moreover, we even identified cases in which the usage of EGO does not provide any benefits over the initial (quasi-)random sample -- especially in case of highly multimodal problems.

Our long-term vision are SMBO approaches that dynamically decide whether to take the next sample from a (quasi-)random distribution or whether to derive it from the surrogate model. Going one step further, we believe that an adaptive choice of the acquisition function, and possibly even of the solver used to optimize the latter, should bring substantial performance gains -- in particular in the case in which the total budget is known in advance. Hence, we need to ``train'' a final recommendation (last evaluation) instead of achieving good anytime performance. These two mentioned questions fall under the umbrella of dynamic algorithm configuration, which has been an important driver for the field of evolutionary computation for the last decades~\cite{BurkeGHKOOQ13,EibenHM99,KarafotiasHE15,DoerrD20chapter}, and which has recently also gained interest in machine learning communities~\cite{BiedenkappBHL19}. 

Typically, the budget of common SMBO applications is too small for a classical a priori (i.e., offline) landscape-aware selection of the optimizer design based on supervised learning approaches (see~\cite{kerschke2018survey} for a survey). However, in case high-level properties -- such as the degree of (multi-)modality or the sizes of the problem's attraction basins -- are known for the problem at hand, or can be guessed by an expert, selecting a suitable initial design strategy is feasible.

Finally, we have seen that the performance of the different designs was often quite comparable. To investigate the differences in more detail, we suggest to consider the different strategies as a portfolio of different algorithms. With this viewpoint, one could analyze the marginal contributions~\cite{XuHHL12} or Shapley values~\cite{FrechetteKMRHL16Shapley} of the different designs, and leverage the information contained therein.

\begin{acks}
This work was supported by the Paris Ile-de-France Region and the \href{https://www.ercis.org}{\textit{European Research Center for Information Systems (ERCIS)}}.
\end{acks}

}%sloppy, pls do not remove

% \bibliographystyle{ACM-Reference-Format}
% \bibliography{smbo}

\begin{thebibliography}{51}

%%% ====================================================================
%%% NOTE TO THE USER: you can override these defaults by providing
%%% customized versions of any of these macros before the \bibliography
%%% command.  Each of them MUST provide its own final punctuation,
%%% except for \shownote{}, \showDOI{}, and \showURL{}.  The latter two
%%% do not use final punctuation, in order to avoid confusing it with
%%% the Web address.
%%%
%%% To suppress output of a particular field, define its macro to expand
%%% to an empty string, or better, \unskip, like this:
%%%
%%% \newcommand{\showDOI}[1]{\unskip}   % LaTeX syntax
%%%
%%% \def \showDOI #1{\unskip}           % plain TeX syntax
%%%
%%% ====================================================================

\ifx \showCODEN    \undefined \def \showCODEN     #1{\unskip}     \fi
\ifx \showDOI      \undefined \def \showDOI       #1{#1}\fi
\ifx \showISBNx    \undefined \def \showISBNx     #1{\unskip}     \fi
\ifx \showISBNxiii \undefined \def \showISBNxiii  #1{\unskip}     \fi
\ifx \showISSN     \undefined \def \showISSN      #1{\unskip}     \fi
\ifx \showLCCN     \undefined \def \showLCCN      #1{\unskip}     \fi
\ifx \shownote     \undefined \def \shownote      #1{#1}          \fi
\ifx \showarticletitle \undefined \def \showarticletitle #1{#1}   \fi
\ifx \showURL      \undefined \def \showURL       {\relax}        \fi
% The following commands are used for tagged output and should be
% invisible to TeX
\providecommand\bibfield[2]{#2}
\providecommand\bibinfo[2]{#2}
\providecommand\natexlab[1]{#1}
\providecommand\showeprint[2][]{arXiv:#2}

\bibitem[\protect\citeauthoryear{Bartz{-}Beielstein}{Bartz{-}Beielstein}{2010}]%
        {SPOT}
\bibfield{author}{\bibinfo{person}{Thomas Bartz{-}Beielstein}.}
  \bibinfo{year}{2010}\natexlab{}.
\newblock \showarticletitle{{SPOT:} An {R} Package For Automatic and
  Interactive Tuning of Optimization Algorithms by Sequential Parameter
  Optimization}.
\newblock \bibinfo{journal}{\emph{CoRR}}  \bibinfo{volume}{abs/1006.4645}
  (\bibinfo{year}{2010}).
\newblock
\showeprint[arxiv]{1006.4645}
\urldef\tempurl%
\url{http://arxiv.org/abs/1006.4645}
\showURL{%
\tempurl}


\bibitem[\protect\citeauthoryear{Bartz-Beielstein and Preuss}{Bartz-Beielstein
  and Preuss}{2006}]%
        {BartzP06}
\bibfield{author}{\bibinfo{person}{Thomas Bartz-Beielstein} {and}
  \bibinfo{person}{Mike Preuss}.} \bibinfo{year}{2006}\natexlab{}.
\newblock \showarticletitle{Considerations of Budget Allocation for Sequential
  Parameter Optimization (SPO)}. In \bibinfo{booktitle}{\emph{Proc. Workshop on
  Empirical Methods for the Analysis of Algorithms (EMAA'06)}}.
  \bibinfo{pages}{35--40}.
\newblock


\bibitem[\protect\citeauthoryear{Beachkofski and Grandhi}{Beachkofski and
  Grandhi}{2002}]%
        {beachkofski_improved_nodate}
\bibfield{author}{\bibinfo{person}{Brian Beachkofski} {and}
  \bibinfo{person}{Ramana Grandhi}.} \bibinfo{year}{2002}\natexlab{}.
\newblock \showarticletitle{Improved {Distributed} {Hypercube} {Sampling}}.
\newblock In \bibinfo{booktitle}{\emph{43rd {AIAA}/{ASME}/{ASCE}/{AHS}/{ASC}
  {Structures}, {Structural} {Dynamics}, and {Materials} {Conference}}}.
  \bibinfo{publisher}{American Institute of Aeronautics and Astronautics}.
\newblock


\bibitem[\protect\citeauthoryear{Belkhir, Dr{\'{e}}o, Sav{\'{e}}ant, and
  Schoenauer}{Belkhir et~al\mbox{.}}{2017}]%
        {BelkhirDSS17}
\bibfield{author}{\bibinfo{person}{Nacim Belkhir}, \bibinfo{person}{Johann
  Dr{\'{e}}o}, \bibinfo{person}{Pierre Sav{\'{e}}ant}, {and}
  \bibinfo{person}{Marc Schoenauer}.} \bibinfo{year}{2017}\natexlab{}.
\newblock \showarticletitle{Per instance algorithm configuration of {CMA-ES}
  with limited budget}. In \bibinfo{booktitle}{\emph{Proc. of Genetic and
  Evolutionary Computation Conference (GECCO'17)}}. \bibinfo{publisher}{ACM},
  \bibinfo{pages}{681--688}.
\newblock


\bibitem[\protect\citeauthoryear{Biedenkapp, Bozkurt, Hutter, and
  Lindauer}{Biedenkapp et~al\mbox{.}}{2019}]%
        {BiedenkappBHL19}
\bibfield{author}{\bibinfo{person}{Andr{\'{e}} Biedenkapp},
  \bibinfo{person}{H.~Furkan Bozkurt}, \bibinfo{person}{Frank Hutter}, {and}
  \bibinfo{person}{Marius Lindauer}.} \bibinfo{year}{2019}\natexlab{}.
\newblock \showarticletitle{Towards White-box Benchmarks for Algorithm
  Control}.
\newblock \bibinfo{journal}{\emph{CoRR}}  \bibinfo{volume}{abs/1906.07644}
  (\bibinfo{year}{2019}).
\newblock
\showeprint[arxiv]{1906.07644}
\urldef\tempurl%
\url{http://arxiv.org/abs/1906.07644}
\showURL{%
\tempurl}


\bibitem[\protect\citeauthoryear{Bischl, Richter, Bossek, Horn, Thomas, and
  Lang}{Bischl et~al\mbox{.}}{2016}]%
        {BRB2017mlrMBO}
\bibfield{author}{\bibinfo{person}{Bernd Bischl}, \bibinfo{person}{Jakob
  Richter}, \bibinfo{person}{Jakob Bossek}, \bibinfo{person}{Daniel Horn},
  \bibinfo{person}{Janek Thomas}, {and} \bibinfo{person}{Michel Lang}.}
  \bibinfo{year}{2016}\natexlab{}.
\newblock \showarticletitle{{mlrMBO: A Modular Framework for Model-Based
  Optimization of Expensive Black-Box Functions}}.
\newblock  (\bibinfo{year}{2016}).
\newblock
\showeprint[arxiv]{stat/1703.03373}
\urldef\tempurl%
\url{http://arxiv.org/abs/1703.03373}
\showURL{%
\tempurl}


\bibitem[\protect\citeauthoryear{Bischl, Wessing, Bauer, Friedrichs, and
  Weihs}{Bischl et~al\mbox{.}}{2014}]%
        {Bischl2014MOIMBO}
\bibfield{author}{\bibinfo{person}{Bernd Bischl}, \bibinfo{person}{Simon
  Wessing}, \bibinfo{person}{Nadja Bauer}, \bibinfo{person}{Klaus Friedrichs},
  {and} \bibinfo{person}{Claus Weihs}.} \bibinfo{year}{2014}\natexlab{}.
\newblock \showarticletitle{MOI-MBO: Multiobjective Infill for Parallel
  Model-Based Optimization}. In \bibinfo{booktitle}{\emph{Learning and
  Intelligent Optimization}}, \bibfield{editor}{\bibinfo{person}{Panos~M.
  Pardalos}, \bibinfo{person}{Mauricio~G.C. Resende},
  \bibinfo{person}{Chrysafis Vogiatzis}, {and} \bibinfo{person}{Jose~L.
  Walteros}} (Eds.). \bibinfo{publisher}{Springer International Publishing},
  \bibinfo{address}{Cham}, \bibinfo{pages}{173--186}.
\newblock
\showISBNx{978-3-319-09584-4}


\bibitem[\protect\citeauthoryear{Bossek}{Bossek}{2017}]%
        {Bossek2017Smoof}
\bibfield{author}{\bibinfo{person}{Jakob Bossek}.}
  \bibinfo{year}{2017}\natexlab{}.
\newblock \showarticletitle{{smoof: Single-and Multi-Objective Optimization
  Test Functions}}.
\newblock \bibinfo{journal}{\emph{The R Journal}} \bibinfo{volume}{9},
  \bibinfo{number}{1} (\bibinfo{year}{2017}), \bibinfo{pages}{103--113}.
\newblock
\urldef\tempurl%
\url{https://journal.r-project.org/archive/2017/RJ-2017-004/RJ-2017-004.pdf}
\showURL{%
\tempurl}


\bibitem[\protect\citeauthoryear{Bossek}{Bossek}{2020a}]%
        {Bossek2020dandy}
\bibfield{author}{\bibinfo{person}{Jakob Bossek}.}
  \bibinfo{year}{2020}\natexlab{a}.
\newblock \bibinfo{booktitle}{\emph{dandy: Designs and Discrepancy}}.
\newblock
\urldef\tempurl%
\url{https://github.com/jakobbossek/dandy}
\showURL{%
\tempurl}
\newblock
\shownote{R package version 1.0.0.0000.}


\bibitem[\protect\citeauthoryear{Bossek}{Bossek}{2020b}]%
        {data}
\bibfield{author}{\bibinfo{person}{Jakob Bossek}.}
  \bibinfo{year}{2020}\natexlab{b}.
\newblock \bibinfo{title}{Public data repository with project data}.
\newblock
\newblock
\urldef\tempurl%
\url{https://github.com/jakobbossek/GECCO2020-smboinitial}
\showURL{%
\tempurl}


\bibitem[\protect\citeauthoryear{Brockhoff, Bischl, and Wagner}{Brockhoff
  et~al\mbox{.}}{2015}]%
        {BrockhoffBW15}
\bibfield{author}{\bibinfo{person}{Dimo Brockhoff}, \bibinfo{person}{Bernd
  Bischl}, {and} \bibinfo{person}{Tobias Wagner}.}
  \bibinfo{year}{2015}\natexlab{}.
\newblock \showarticletitle{The Impact of Initial Designs on the Performance of
  MATSuMoTo on the Noiseless {BBOB-2015} Testbed: {A} Preliminary Study}. In
  \bibinfo{booktitle}{\emph{Proc. of Genetic and Evolutionary Computation
  Conference (GECCO'15)}}. \bibinfo{publisher}{ACM},
  \bibinfo{pages}{1159--1166}.
\newblock


\bibitem[\protect\citeauthoryear{Burke, Gendreau, Hyde, Kendall, Ochoa,
  {\"{O}}zcan, and Qu}{Burke et~al\mbox{.}}{2013}]%
        {BurkeGHKOOQ13}
\bibfield{author}{\bibinfo{person}{Edmund~K. Burke}, \bibinfo{person}{Michel
  Gendreau}, \bibinfo{person}{Matthew~R. Hyde}, \bibinfo{person}{Graham
  Kendall}, \bibinfo{person}{Gabriela Ochoa}, \bibinfo{person}{Ender
  {\"{O}}zcan}, {and} \bibinfo{person}{Rong Qu}.}
  \bibinfo{year}{2013}\natexlab{}.
\newblock \showarticletitle{Hyper-heuristics: a survey of the state of the
  art}.
\newblock \bibinfo{journal}{\emph{{JORS}}} \bibinfo{volume}{64},
  \bibinfo{number}{12} (\bibinfo{year}{2013}), \bibinfo{pages}{1695--1724}.
\newblock


\bibitem[\protect\citeauthoryear{Carnell}{Carnell}{2019}]%
        {Rlhs}
\bibfield{author}{\bibinfo{person}{Rob Carnell}.}
  \bibinfo{year}{2019}\natexlab{}.
\newblock \bibinfo{booktitle}{\emph{lhs: Latin Hypercube Samples}}.
\newblock
\urldef\tempurl%
\url{https://CRAN.R-project.org/package=lhs}
\showURL{%
\tempurl}
\newblock
\shownote{R package version 1.0.1.}


\bibitem[\protect\citeauthoryear{Christophe and Petr}{Christophe and
  Petr}{2019}]%
        {Rrndtoolbox}
\bibfield{author}{\bibinfo{person}{Dutang Christophe} {and}
  \bibinfo{person}{Savicky Petr}.} \bibinfo{year}{2019}\natexlab{}.
\newblock \bibinfo{booktitle}{\emph{randtoolbox: Generating and Testing Random
  Numbers}}.
\newblock
\newblock
\shownote{R package version 1.30.0.}


\bibitem[\protect\citeauthoryear{Dick and Pillichshammer}{Dick and
  Pillichshammer}{2010}]%
        {DickP10}
\bibfield{author}{\bibinfo{person}{Josef Dick} {and} \bibinfo{person}{Friedrich
  Pillichshammer}.} \bibinfo{year}{2010}\natexlab{}.
\newblock \bibinfo{booktitle}{\emph{Digital Nets and Sequences}}.
\newblock \bibinfo{publisher}{Cambridge University Press}.
\newblock


\bibitem[\protect\citeauthoryear{Doerr and Doerr}{Doerr and Doerr}{2020}]%
        {DoerrD20chapter}
\bibfield{author}{\bibinfo{person}{Benjamin Doerr} {and}
  \bibinfo{person}{Carola Doerr}.} \bibinfo{year}{2020}\natexlab{}.
\newblock \showarticletitle{Theory of Parameter Control for Discrete Black-Box
  Optimization: Provable Performance Gains Through Dynamic Parameter Choices}.
\newblock In \bibinfo{booktitle}{\emph{Theory of Evolutionary Computation:
  Recent Developments in Discrete Optimization}}.
  \bibinfo{publisher}{Springer}, \bibinfo{pages}{271--321}.
\newblock


\bibitem[\protect\citeauthoryear{Eiben, Hinterding, and Michalewicz}{Eiben
  et~al\mbox{.}}{1999}]%
        {EibenHM99}
\bibfield{author}{\bibinfo{person}{{\'A}goston~Endre Eiben},
  \bibinfo{person}{Robert Hinterding}, {and} \bibinfo{person}{Zbigniew
  Michalewicz}.} \bibinfo{year}{1999}\natexlab{}.
\newblock \showarticletitle{Parameter control in evolutionary algorithms}.
\newblock \bibinfo{journal}{\emph{IEEE Transactions on Evolutionary
  Computation}}  \bibinfo{volume}{3} (\bibinfo{year}{1999}),
  \bibinfo{pages}{124--141}.
\newblock


\bibitem[\protect\citeauthoryear{Falkner, Klein, and Hutter}{Falkner
  et~al\mbox{.}}{2018}]%
        {BOHB}
\bibfield{author}{\bibinfo{person}{Stefan Falkner}, \bibinfo{person}{Aaron
  Klein}, {and} \bibinfo{person}{Frank Hutter}.}
  \bibinfo{year}{2018}\natexlab{}.
\newblock \showarticletitle{{BOHB}: Robust and Efficient Hyperparameter
  Optimization at Scale}. In \bibinfo{booktitle}{\emph{ICML}}.
  \bibinfo{pages}{1436--1445}.
\newblock


\bibitem[\protect\citeauthoryear{Faure and Tezuka}{Faure and Tezuka}{2002}]%
        {FaureT2002}
\bibfield{author}{\bibinfo{person}{Henri Faure} {and} \bibinfo{person}{Shu
  Tezuka}.} \bibinfo{year}{2002}\natexlab{}.
\newblock \showarticletitle{Another Random Scrambling of Digital
  (t,s)-Sequences}. In \bibinfo{booktitle}{\emph{Monte Carlo and Quasi-Monte
  Carlo Methods 2000}}, \bibfield{editor}{\bibinfo{person}{Kai-Tai Fang},
  \bibinfo{person}{Harald Niederreiter}, {and} \bibinfo{person}{Fred~J.
  Hickernell}} (Eds.). \bibinfo{publisher}{Springer Berlin Heidelberg},
  \bibinfo{address}{Berlin, Heidelberg}, \bibinfo{pages}{242--256}.
\newblock
\showISBNx{978-3-642-56046-0}


\bibitem[\protect\citeauthoryear{Fr{\'{e}}chette, Kotthoff, Michalak, Rahwan,
  Hoos, and Leyton{-}Brown}{Fr{\'{e}}chette et~al\mbox{.}}{2016}]%
        {FrechetteKMRHL16Shapley}
\bibfield{author}{\bibinfo{person}{Alexandre Fr{\'{e}}chette},
  \bibinfo{person}{Lars Kotthoff}, \bibinfo{person}{Tomasz~P. Michalak},
  \bibinfo{person}{Talal Rahwan}, \bibinfo{person}{Holger~H. Hoos}, {and}
  \bibinfo{person}{Kevin Leyton{-}Brown}.} \bibinfo{year}{2016}\natexlab{}.
\newblock \showarticletitle{Using the Shapley Value to Analyze Algorithm
  Portfolios}. In \bibinfo{booktitle}{\emph{Proceedings of the Thirtieth {AAAI}
  Conference on Artificial Intelligence, February 12-17, 2016, Phoenix,
  Arizona, {USA}}}. \bibinfo{publisher}{AAAI}, \bibinfo{pages}{3397--3403}.
\newblock


\bibitem[\protect\citeauthoryear{Halton}{Halton}{1960}]%
        {Halton60}
\bibfield{author}{\bibinfo{person}{John~H. Halton}.}
  \bibinfo{year}{1960}\natexlab{}.
\newblock \showarticletitle{On the efficiency of certain quasi-random sequences
  of points in evaluating multi-dimensional integrals}.
\newblock \bibinfo{journal}{\emph{Numer. Math.}}  \bibinfo{volume}{2}
  (\bibinfo{year}{1960}), \bibinfo{pages}{84--90}.
\newblock


\bibitem[\protect\citeauthoryear{Hansen, Auger, Mersmann, Tu{\v s}ar, and
  Brockhoff}{Hansen et~al\mbox{.}}{2016}]%
        {hansen2016cocoplat}
\bibfield{author}{\bibinfo{person}{Nikolaus Hansen}, \bibinfo{person}{Anne
  Auger}, \bibinfo{person}{Olaf Mersmann}, \bibinfo{person}{Tea Tu{\v s}ar},
  {and} \bibinfo{person}{Dimo Brockhoff}.} \bibinfo{year}{2016}\natexlab{}.
\newblock \showarticletitle{{COCO: A Platform for Comparing Continuous
  Optimizers in a Black-Box Setting}}.
\newblock \bibinfo{journal}{\emph{ArXiv e-prints}}
  \bibinfo{volume}{arXiv:1603.08785} (\bibinfo{year}{2016}).
\newblock


\bibitem[\protect\citeauthoryear{Hansen, Finck, Ros, and Auger}{Hansen
  et~al\mbox{.}}{2009}]%
        {Hansen2009_Noiseless}
\bibfield{author}{\bibinfo{person}{Nikolaus Hansen}, \bibinfo{person}{Steffen
  Finck}, \bibinfo{person}{Raymond Ros}, {and} \bibinfo{person}{Anne Auger}.}
  \bibinfo{year}{2009}\natexlab{}.
\newblock \bibinfo{booktitle}{\emph{{Real-Parameter Black-Box Optimization
  Benchmarking 2009: Noiseless Functions Definitions}}}.
\newblock \bibinfo{type}{{T}echnical {R}eport} RR-6829.
  \bibinfo{institution}{{INRIA}}.
\newblock
\urldef\tempurl%
\url{https://hal.inria.fr/inria-00362633/document}
\showURL{%
\tempurl}


\bibitem[\protect\citeauthoryear{Hansen and Ostermeier}{Hansen and
  Ostermeier}{2001}]%
        {HansenO01}
\bibfield{author}{\bibinfo{person}{Nikolaus Hansen} {and}
  \bibinfo{person}{Andreas Ostermeier}.} \bibinfo{year}{2001}\natexlab{}.
\newblock \showarticletitle{Completely Derandomized Self-Adaptation in
  Evolution Strategies}.
\newblock \bibinfo{journal}{\emph{Evol.~Computation}} \bibinfo{volume}{9},
  \bibinfo{number}{2} (\bibinfo{year}{2001}), \bibinfo{pages}{159--195}.
\newblock


\bibitem[\protect\citeauthoryear{Haziza, Rapin, and Synnaeve}{Haziza
  et~al\mbox{.}}{2020}]%
        {HiPlot}
\bibfield{author}{\bibinfo{person}{Daniel Haziza},
  \bibinfo{person}{J{\'e}r{\'e}my Rapin}, {and} \bibinfo{person}{Gabriel
  Synnaeve}.} \bibinfo{year}{2020}\natexlab{}.
\newblock \bibinfo{title}{{HiPlot - High dimensional Interactive Plotting}}.
\newblock
  \bibinfo{howpublished}{\url{https://github.com/facebookresearch/hiplot}}.
\newblock


\bibitem[\protect\citeauthoryear{Hofert and Lemieux}{Hofert and
  Lemieux}{2019}]%
        {Rqrng}
\bibfield{author}{\bibinfo{person}{Marius Hofert} {and}
  \bibinfo{person}{Christiane Lemieux}.} \bibinfo{year}{2019}\natexlab{}.
\newblock \bibinfo{booktitle}{\emph{qrng: (Randomized) Quasi-Random Number
  Generators}}.
\newblock
\urldef\tempurl%
\url{https://CRAN.R-project.org/package=qrng}
\showURL{%
\tempurl}
\newblock
\shownote{R package version 0.0-7.}


\bibitem[\protect\citeauthoryear{Horn, Wagner, Biermann, Weihs, and
  Bischl}{Horn et~al\mbox{.}}{2015}]%
        {HWBWB2015}
\bibfield{author}{\bibinfo{person}{Daniel Horn}, \bibinfo{person}{Tobias
  Wagner}, \bibinfo{person}{Dirk Biermann}, \bibinfo{person}{Claus Weihs},
  {and} \bibinfo{person}{Bernd Bischl}.} \bibinfo{year}{2015}\natexlab{}.
\newblock \showarticletitle{Model-Based Multi-objective Optimization: Taxonomy,
  Multi-Point Proposal, Toolbox and Benchmark}. In
  \bibinfo{booktitle}{\emph{Evolutionary Multi-Criterion Optimization}}.
  \bibinfo{publisher}{Springer International Publishing},
  \bibinfo{address}{Cham}, \bibinfo{pages}{64--78}.
\newblock
\showISBNx{978-3-319-15934-8}


\bibitem[\protect\citeauthoryear{Hutter, Hoos, and Leyton-Brown}{Hutter
  et~al\mbox{.}}{2011}]%
        {SMAC}
\bibfield{author}{\bibinfo{person}{Frank Hutter}, \bibinfo{person}{Holger~H.
  Hoos}, {and} \bibinfo{person}{Kevin Leyton-Brown}.}
  \bibinfo{year}{2011}\natexlab{}.
\newblock \showarticletitle{Sequential model-based optimization for general
  algorithm configuration}. In \bibinfo{booktitle}{\emph{LION}}. Springer,
  \bibinfo{pages}{507--523}.
\newblock


\bibitem[\protect\citeauthoryear{Jones}{Jones}{2001}]%
        {Jones2001}
\bibfield{author}{\bibinfo{person}{Donald~R. Jones}.}
  \bibinfo{year}{2001}\natexlab{}.
\newblock \showarticletitle{A Taxonomy of Global Optimization Methods Based on
  Response Surfaces}.
\newblock \bibinfo{journal}{\emph{Journal of Global Optimization}}
  \bibinfo{volume}{21}, \bibinfo{number}{4} (\bibinfo{date}{01 Dec}
  \bibinfo{year}{2001}), \bibinfo{pages}{345--383}.
\newblock
\showISSN{1573-2916}


\bibitem[\protect\citeauthoryear{Jones, Schonlau, and Welch}{Jones
  et~al\mbox{.}}{1998}]%
        {JonesSW98}
\bibfield{author}{\bibinfo{person}{Donald~R. Jones}, \bibinfo{person}{Matthias
  Schonlau}, {and} \bibinfo{person}{William~J. Welch}.}
  \bibinfo{year}{1998}\natexlab{}.
\newblock \showarticletitle{Efficient Global Optimization of Expensive
  Black-Box Functions}.
\newblock \bibinfo{journal}{\emph{Journal of Global Optimization}}
  \bibinfo{volume}{13}, \bibinfo{number}{4} (\bibinfo{year}{1998}),
  \bibinfo{pages}{455--492}.
\newblock


\bibitem[\protect\citeauthoryear{Karafotias, Hoogendoorn, and Eiben}{Karafotias
  et~al\mbox{.}}{2015}]%
        {KarafotiasHE15}
\bibfield{author}{\bibinfo{person}{Giorgos Karafotias}, \bibinfo{person}{Mark
  Hoogendoorn}, {and} \bibinfo{person}{{\'A}goston~Endre Eiben}.}
  \bibinfo{year}{2015}\natexlab{}.
\newblock \showarticletitle{Parameter Control in Evolutionary Algorithms:
  Trends and Challenges}.
\newblock \bibinfo{journal}{\emph{IEEE Transactions on Evolutionary
  Computation}}  \bibinfo{volume}{19} (\bibinfo{year}{2015}),
  \bibinfo{pages}{167--187}.
\newblock


\bibitem[\protect\citeauthoryear{Kerschke, Hoos, Neumann, and
  Trautmann}{Kerschke et~al\mbox{.}}{2019}]%
        {kerschke2018survey}
\bibfield{author}{\bibinfo{person}{Pascal Kerschke}, \bibinfo{person}{Holger~H.
  Hoos}, \bibinfo{person}{Frank Neumann}, {and} \bibinfo{person}{Heike
  Trautmann}.} \bibinfo{year}{2019}\natexlab{}.
\newblock \showarticletitle{Automated Algorithm Selection: Survey and
  Perspectives}.
\newblock \bibinfo{journal}{\emph{Evolutionary Computation}}
  \bibinfo{volume}{27}, \bibinfo{number}{1} (\bibinfo{year}{2019}),
  \bibinfo{pages}{3--45}.
\newblock


\bibitem[\protect\citeauthoryear{Kerschke and Trautmann}{Kerschke and
  Trautmann}{2019}]%
        {KerschkeT2019AutomatedAlgorithm}
\bibfield{author}{\bibinfo{person}{Pascal Kerschke} {and}
  \bibinfo{person}{Heike Trautmann}.} \bibinfo{year}{2019}\natexlab{}.
\newblock \showarticletitle{{Automated Algorithm Selection on Continuous
  Black-Box Problems By Combining Exploratory Landscape Analysis and Machine
  Learning}}.
\newblock \bibinfo{journal}{\emph{Evolutionary Computation (ECJ)}}
  \bibinfo{volume}{27}, \bibinfo{number}{1} (\bibinfo{year}{2019}),
  \bibinfo{pages}{99~--~127}.
\newblock


\bibitem[\protect\citeauthoryear{Knowles}{Knowles}{2006}]%
        {Knowles2006ParEGO}
\bibfield{author}{\bibinfo{person}{Joshua Knowles}.}
  \bibinfo{year}{2006}\natexlab{}.
\newblock \showarticletitle{ParEGO: a hybrid algorithm with on-line landscape
  approximation for expensive multiobjective optimization problems}.
\newblock \bibinfo{journal}{\emph{IEEE Transactions on Evolutionary
  Computation}}  \bibinfo{volume}{10} (\bibinfo{year}{2006}),
  \bibinfo{pages}{50--66}.
\newblock


\bibitem[\protect\citeauthoryear{Kotthoff, Thornton, Hoos, Hutter, and
  Leyton{-}Brown}{Kotthoff et~al\mbox{.}}{2019}]%
        {KotthoffTHHL19}
\bibfield{author}{\bibinfo{person}{Lars Kotthoff}, \bibinfo{person}{Chris
  Thornton}, \bibinfo{person}{Holger~H. Hoos}, \bibinfo{person}{Frank Hutter},
  {and} \bibinfo{person}{Kevin Leyton{-}Brown}.}
  \bibinfo{year}{2019}\natexlab{}.
\newblock \showarticletitle{Auto-WEKA: Automatic Model Selection and
  Hyperparameter Optimization in {WEKA}}.
\newblock In \bibinfo{booktitle}{\emph{Automated Machine Learning - Methods,
  Systems, Challenges}}. \bibinfo{publisher}{Springer},
  \bibinfo{pages}{81--95}.
\newblock


\bibitem[\protect\citeauthoryear{Lindauer, Feurer, Eggensperger, Biedenkapp,
  and Hutter}{Lindauer et~al\mbox{.}}{2019}]%
        {Lindauer19}
\bibfield{author}{\bibinfo{person}{Marius Lindauer}, \bibinfo{person}{Matthias
  Feurer}, \bibinfo{person}{Katharina Eggensperger},
  \bibinfo{person}{Andr{\'{e}} Biedenkapp}, {and} \bibinfo{person}{Frank
  Hutter}.} \bibinfo{year}{2019}\natexlab{}.
\newblock \showarticletitle{Towards Assessing the Impact of Bayesian
  Optimization's Own Hyperparameters}. In \bibinfo{booktitle}{\emph{{IJCAI}
  2019 {DSO} Workshop}}.
\newblock


\bibitem[\protect\citeauthoryear{Lobo, Lima, and Michalewicz}{Lobo
  et~al\mbox{.}}{2007}]%
        {LoboLM07}
\bibfield{editor}{\bibinfo{person}{Fernando~G. Lobo},
  \bibinfo{person}{Cl{\'{a}}udio~F. Lima}, {and} \bibinfo{person}{Zbigniew
  Michalewicz}} (Eds.). \bibinfo{year}{2007}\natexlab{}.
\newblock \bibinfo{booktitle}{\emph{Parameter Setting in Evolutionary
  Algorithms}}. \bibinfo{series}{Studies in Computational Intelligence},
  Vol.~\bibinfo{volume}{54}.
\newblock \bibinfo{publisher}{Springer}.
\newblock
\showISBNx{978-3-540-69431-1}


\bibitem[\protect\citeauthoryear{Matsumoto and Nishimura}{Matsumoto and
  Nishimura}{1998}]%
        {MersenneTwister}
\bibfield{author}{\bibinfo{person}{Makoto Matsumoto} {and}
  \bibinfo{person}{Takuji Nishimura}.} \bibinfo{year}{1998}\natexlab{}.
\newblock \showarticletitle{Mersenne Twister: A 623-Dimensionally
  Equidistributed Uniform Pseudo-Random Number Generator}.
\newblock \bibinfo{journal}{\emph{ACM Trans. Model. Comput. Simul.}}
  \bibinfo{volume}{8}, \bibinfo{number}{1} (\bibinfo{date}{Jan.}
  \bibinfo{year}{1998}), \bibinfo{pages}{3–30}.
\newblock
\showISSN{1049-3301}


\bibitem[\protect\citeauthoryear{McKay, Beckman, and Conover}{McKay
  et~al\mbox{.}}{1979}]%
        {LHS}
\bibfield{author}{\bibinfo{person}{Michael~D. McKay},
  \bibinfo{person}{Richard~J. Beckman}, {and} \bibinfo{person}{William~J.
  Conover}.} \bibinfo{year}{1979}\natexlab{}.
\newblock \showarticletitle{{A Comparison of Three Methods for Selecting Values
  of Input Variables in the Analysis of Output from a Computer Code}}.
\newblock \bibinfo{journal}{\emph{Technometrics}}  \bibinfo{volume}{21}
  (\bibinfo{year}{1979}), \bibinfo{pages}{239--245}.
\newblock
\showISSN{00401706}


\bibitem[\protect\citeauthoryear{Mockus}{Mockus}{1989}]%
        {MockusBO}
\bibfield{editor}{\bibinfo{person}{Jonas Mockus}} (Ed.).
  \bibinfo{year}{1989}\natexlab{}.
\newblock \bibinfo{booktitle}{\emph{Bayesian Approach to Global Optimization}}.
\newblock \bibinfo{publisher}{Springer}.
\newblock
\showISBNx{978-94-009-0909-0}


\bibitem[\protect\citeauthoryear{Morar, Knowles, and Sampaio}{Morar
  et~al\mbox{.}}{2017}]%
        {MorarKS17}
\bibfield{author}{\bibinfo{person}{Marius~Tudor Morar}, \bibinfo{person}{Joshua
  Knowles}, {and} \bibinfo{person}{Sandra Sampaio}.}
  \bibinfo{year}{2017}\natexlab{}.
\newblock \showarticletitle{Initialization of {B}ayesian Optimization Viewed as
  Part of a Larger Algorithm Portfolio}. In \bibinfo{booktitle}{\emph{Proc. of
  the international workshop in Data Science meets Optimization (DSO at {CEC
  and CPAIOR} 2017)}}.
\newblock


\bibitem[\protect\citeauthoryear{Mueller}{Mueller}{2014}]%
        {mueller2014matsumoto}
\bibfield{author}{\bibinfo{person}{Juliane Mueller}.}
  \bibinfo{year}{2014}\natexlab{}.
\newblock \bibinfo{title}{MATSuMoTo: The MATLAB Surrogate Model Toolbox For
  Computationally Expensive Black-Box Global Optimization Problems}.
\newblock
\newblock
\showeprint[arxiv]{math.OC/1404.4261}


\bibitem[\protect\citeauthoryear{Nelder and Mead}{Nelder and Mead}{1965}]%
        {NelderM65}
\bibfield{author}{\bibinfo{person}{John~Ashworth Nelder} {and}
  \bibinfo{person}{Roger Mead}.} \bibinfo{year}{1965}\natexlab{}.
\newblock \showarticletitle{{{A Simplex Method for Function Minimization}}}.
\newblock \bibinfo{journal}{\emph{Comput. J.}}  \bibinfo{volume}{7}
  (\bibinfo{year}{1965}), \bibinfo{pages}{308--313}.
\newblock
\showISSN{0010-4620}


\bibitem[\protect\citeauthoryear{Owen}{Owen}{1995}]%
        {Owen1995}
\bibfield{author}{\bibinfo{person}{Art~B. Owen}.}
  \bibinfo{year}{1995}\natexlab{}.
\newblock \showarticletitle{Randomly Permuted (t,m,s)-Nets and (t,
  s)-Sequences}. In \bibinfo{booktitle}{\emph{Monte Carlo and Quasi-Monte Carlo
  Methods in Scientific Computing}}, \bibfield{editor}{\bibinfo{person}{Harald
  Niederreiter} {and} \bibinfo{person}{Peter Jau-Shyong Shiue}} (Eds.).
  \bibinfo{publisher}{Springer New York}, \bibinfo{address}{New York, NY},
  \bibinfo{pages}{299--317}.
\newblock
\showISBNx{978-1-4612-2552-2}


\bibitem[\protect\citeauthoryear{{R Core Team}}{{R Core Team}}{2018}]%
        {Rcore2018}
\bibfield{author}{\bibinfo{person}{{R Core Team}}.}
  \bibinfo{year}{2018}\natexlab{}.
\newblock \bibinfo{booktitle}{\emph{R: A Language and Environment for
  Statistical Computing}}.
\newblock R Foundation for Statistical Computing, Vienna, Austria.
\newblock
\urldef\tempurl%
\url{https://www.R-project.org/}
\showURL{%
\tempurl}


\bibitem[\protect\citeauthoryear{Rasmussen and Williams}{Rasmussen and
  Williams}{2006}]%
        {GaussianPML}
\bibfield{editor}{\bibinfo{person}{Carl~Edward Rasmussen} {and}
  \bibinfo{person}{Christopher K.~I. Williams}} (Eds.).
  \bibinfo{year}{2006}\natexlab{}.
\newblock \bibinfo{booktitle}{\emph{Gaussian Processes for Machine Learning}}.
\newblock \bibinfo{publisher}{The MIT Press}.
\newblock
\showISBNx{0-262-18253-X}


\bibitem[\protect\citeauthoryear{Saini, L{\'o}pez-Ib{\'a}{\~{n}}ez, and
  Miettinen}{Saini et~al\mbox{.}}{2019}]%
        {saini2019automatic}
\bibfield{author}{\bibinfo{person}{Bhupinder~Singh Saini},
  \bibinfo{person}{Manuel L{\'o}pez-Ib{\'a}{\~{n}}ez}, {and}
  \bibinfo{person}{Kaisa Miettinen}.} \bibinfo{year}{2019}\natexlab{}.
\newblock \showarticletitle{{Automatic Surrogate Modelling Technique Selection
  Based on Features of Optimization Problems}}. In
  \bibinfo{booktitle}{\emph{Proceedings of the Genetic and Evolutionary
  Computation Conference (GECCO) Companion}}. \bibinfo{publisher}{ACM},
  \bibinfo{pages}{1765~--~1772}.
\newblock


\bibitem[\protect\citeauthoryear{Santner, Williams, and Notz}{Santner
  et~al\mbox{.}}{2003}]%
        {santner_design_2003}
\bibfield{author}{\bibinfo{person}{T.J. Santner}, \bibinfo{person}{B.J.
  Williams}, {and} \bibinfo{person}{W.I. Notz}.}
  \bibinfo{year}{2003}\natexlab{}.
\newblock \bibinfo{booktitle}{\emph{The {Design} and {Analysis} of {Computer}
  {Experiments}}}.
\newblock \bibinfo{publisher}{Springer}.
\newblock
\showISBNx{978-1-4419-2992-1}


\bibitem[\protect\citeauthoryear{Shahriari, Swersky, Wang, Adams, and
  de~Freitas}{Shahriari et~al\mbox{.}}{2016}]%
        {ShahriariSWAF16}
\bibfield{author}{\bibinfo{person}{Bobak Shahriari}, \bibinfo{person}{Kevin
  Swersky}, \bibinfo{person}{Ziyu Wang}, \bibinfo{person}{Ryan~P. Adams}, {and}
  \bibinfo{person}{Nando de Freitas}.} \bibinfo{year}{2016}\natexlab{}.
\newblock \showarticletitle{Taking the Human Out of the Loop: {A} Review of
  Bayesian Optimization}.
\newblock \bibinfo{journal}{\emph{Proc. IEEE}} \bibinfo{volume}{104},
  \bibinfo{number}{1} (\bibinfo{year}{2016}), \bibinfo{pages}{148--175}.
\newblock


\bibitem[\protect\citeauthoryear{Sobol}{Sobol}{1967}]%
        {Sobol}
\bibfield{author}{\bibinfo{person}{Ilya~Meyerovich Sobol}.}
  \bibinfo{year}{1967}\natexlab{}.
\newblock \showarticletitle{On the distribution of points in a cube and the
  approximate evaluation of integrals}.
\newblock \bibinfo{journal}{\emph{U. S. S. R. Comput. Math. and Math. Phys.}}
  \bibinfo{volume}{7}, \bibinfo{number}{4} (\bibinfo{date}{Jan.}
  \bibinfo{year}{1967}), \bibinfo{pages}{86--112}.
\newblock
\showISSN{00415553}


\bibitem[\protect\citeauthoryear{Xu, Hutter, Hoos, and Leyton{-}Brown}{Xu
  et~al\mbox{.}}{2012}]%
        {XuHHL12}
\bibfield{author}{\bibinfo{person}{Lin Xu}, \bibinfo{person}{Frank Hutter},
  \bibinfo{person}{Holger~H. Hoos}, {and} \bibinfo{person}{Kevin
  Leyton{-}Brown}.} \bibinfo{year}{2012}\natexlab{}.
\newblock \showarticletitle{Evaluating Component Solver Contributions to
  Portfolio-Based Algorithm Selectors}. In \bibinfo{booktitle}{\emph{Proc. of
  Theory and Applications of Satisfiability Testing (SAT'12)}}
  \emph{(\bibinfo{series}{Lecture Notes in Computer Science})},
  Vol.~\bibinfo{volume}{7317}. \bibinfo{publisher}{Springer},
  \bibinfo{pages}{228--241}.
\newblock


\end{thebibliography}
%%% -*-BibTeX-*-
%%% Do NOT edit. File created by BibTeX with style
%%% ACM-Reference-Format-Journals [18-Jan-2012].

\end{document}